\documentclass{article} 
\PassOptionsToPackage{table,xcdraw}{xcolor}
\usepackage[final]{colm2025_conference}

\usepackage{microtype}
\usepackage{hyperref}
\usepackage{url}
\usepackage{booktabs}
\usepackage{amsmath}
\usepackage{graphicx}
\usepackage{multirow}
\usepackage{array}
\usepackage{amsmath, amssymb, amsthm}
\newtheorem{theorem}{Theorem}
\usepackage{lineno}
\usepackage{algorithm}
\usepackage{algorithmic}
\usepackage{subcaption}

\definecolor{darkblue}{rgb}{0, 0, 0.5}
\hypersetup{colorlinks=true, citecolor=darkblue, linkcolor=darkblue, urlcolor=darkblue}

\title{Improving Fisher Information Estimation and Efficiency \\ for LoRA-based LLM Unlearning}

\author{Yejin Kim\thanks{Co-first authors. \quad $\dag$ Co-corresponding authors.} \\
Sogang University\\
\And
Eunwon Kim\footnotemark[1]~ \\
Sogang University\\
\And
Buru Chang\dag \\
Korea University \\
\And
Junsuk Choe\dag \\
Sogang University \\
}

\begin{document}

\ifcolmsubmission
\linenumbers
\fi

\maketitle

\begin{abstract}

LLMs have demonstrated remarkable performance across various tasks but face challenges related to unintentionally generating outputs containing sensitive information. A straightforward approach to address this issue is to retrain the model after excluding the problematic data. However, this approach incurs prohibitively high computational costs. To overcome this limitation, machine unlearning has emerged as a promising solution that can effectively remove sensitive information without the need to retrain the model from scratch. Recently, FILA has been proposed as a parameter-efficient unlearning method by integrating LoRA adapters. Specifically, it calculates the Fisher information to identify parameters associated with the forget set and assigns them to LoRA adapters for updates. Despite its innovative approach, FILA still requires access to all model parameters and does not adequately account for fundamental assumptions underlying Fisher information, leading to inaccuracies in importance estimation. To address these limitations, we propose VILA, a novel unlearning framework that explicitly considers the assumptions overlooked in FILA, thereby enhancing the accuracy of parameter identification for the forget set. Moreover, VILA significantly reduces computational costs by enabling parameter identification without accessing the entire model. Our method achieves up to 100× higher parameter efficiency and 40× faster training speed compared to FILA, and sets new state-of-the-art performance on benchmarks including TOFU, WMDP, and MUSE. Our code is available at \url{https://github.com/kyj93790/VILA}.
\end{abstract}
\section{Introduction}

Large Language Models (LLMs) are driving remarkable progress across a wide range of applications. However, they also exhibit a critical risk: the tendency to memorize and regenerate sensitive personal information or copyrighted content from their training data. For instance, \cite{brown2022privacy} have shown that LLMs often output personal identifiers such as email addresses and phone numbers from the training corpus. Similarly, LLMs are known to reproduce copyrighted materials, such as passages from Harry Potter, with high fidelity \citep{eldan2023whos}. These issues raise serious concerns about privacy violations and intellectual property infringement. As a result, there is growing demand for methods that can effectively remove sensitive or proprietary information from LLMs.

The most straightforward way to remove specific information from a model is to retrain it from scratch without the corresponding data (\textit{i.e.}, exact unlearning). However, given the massive size of LLMs and their extensive training corpora, this approach is computationally expensive and time-consuming. To address this challenge, recent research has focused on methods that aim to eliminate the information to be forgotten without full retraining (\textit{i.e.}, approximate unlearning). For example, loss-based techniques such as Gradient Ascent (GA)~\citep{jang2023knowledge}, Negative Preference Optimization (NPO)~\citep{zhang2024negative}, and Inverted Hinge Loss (IHL)~\citep{cha2025fila} have been proposed to reduce the likelihood of generating specific content through fine-tuning. 

Nevertheless, directly updating billions of parameters remains computationally demanding, even when applying approximate unlearning techniques. To alleviate this burden, Fisher-Initialization of Low-rank Adapters (FILA)~\citep{cha2025fila} has been introduced. FILA leverages Fisher information~\citep{fisher1922mathematical} to estimate gradient variance and identify parameters most closely related to the data to be forgotten. These parameters are isolated from the base model by assigning them to a LoRA adapter~\citep{hu2022lora}. Unlearning is then performed exclusively on the adapters. This enables parameter-efficient unlearning while minimizing the impact on the retained knowledge.

However, our analysis reveals two critical limitations of FILA. First, the Fisher information used by FILA does not accurately represent parameter importance in the machine unlearning setting. For Fisher information to indicate importance, the distribution of the forget set must match that of the full dataset. However, the forget set typically constitutes only a small fraction of the entire dataset, inevitably leading to a statistical discrepancy between the forget set and the full dataset. FILA overlooks this discrepancy, which results in a forget importance map that inaccurately captures the association between the forget set and the parameters. Moreover, although FILA is designed for parameter-efficient unlearning, it still requires computing full gradients for all model parameters to construct the importance map. This significantly undermines its computational efficiency. Our analysis shows that the cost of FILA grows rapidly with the size of the forget set. When forgetting 10\% of the dataset, the initialization time exceeds that of full model retraining—highlighting a serious limitation in scalability (refer to Section~\ref{subsec:improved_efficiency}).

Building on the above analysis, we propose a precise and scalable approach, Variance-based Importance estimation and efficient Low-rank Adaptation (\textbf{VILA}). Our method improves the estimation of parameter importance by explicitly considering the distributional shift of the forget set. Furthermore, we construct the forget importance map solely using the gradients from the LoRA adapters, resulting in up to a 40× speedup and approximately 100× reduction in memory consumption compared to FILA as the size of the forget set increases.

We evaluate our method on multiple LLMs, including Phi-1.5B~\citep{li2023phi15}, Llama2-7B~\citep{touvron2023llama}, Zephyr-7B~\citep{tunstall2024zephyr} and ICLM-7B~\citep{shi2024incontext}, in combination with existing unlearning loss functions such as GA, NPO, and IHL. Experimental results on the TOFU~\citep{maini2024tofu}, MUSE Books~\citep{shi2025muse}, WMDP Bio and WMDP Cyber~\citep{li2024wmdp} benchmarks demonstrate that our method not only improves resource efficiency but also sets a new state-of-the-art in unlearning performance.
\section{Related Work}\label{sec:2_related_work}
LLM Unlearning aims to eliminate the influence of specific data from large language models without incurring the cost of expensive retraining. This approach addresses various challenges, such as preserving privacy, resolving copyright issues, and removing hazardous knowledge~\citep{brown2022privacy, eldan2023whos, li2024wmdp}.

Several studies mainly focus on modifying the loss function to induce unlearning. A representative example is Gradient Ascent (GA), which increases the loss on the forget data in order to reduce the model’s predictive accuracy on that data~\citep{jang2023knowledge}. The limitation of GA is that it can easily degrade performance on retain data~\citep{maini2024tofu}. To address this issue, Gradient Difference (GD) has been introduced, performing gradient ascent on forget data to eliminate their influence while applying gradient descent on retain data to preserve the model’s generalization ability~\citep{liu2022continual}. Also, Negative Preference Optimization (NPO)~\citep{zhang2024negative} has been proposed, building on the LLM alignment approach~\citep{rafailov2024direct}. By reweighting gradients during the learning process, NPO addresses the issue of excessive unlearning commonly caused by GA, significantly improving the stability of the unlearning process. Most recently, Inverted Hinge Loss (IHL)~\citep{cha2025fila} promotes unlearning by decreasing the probability of the forget token while increasing the probability of the highest-probability alternative token, excluding the forget token itself.

Beyond loss function-based methods, various approaches have been proposed. Task Arithmetic~\citep{ilharco2023editing} defines the difference between the fine-tuned model only on the forget set and the original model as a \textit{task vector}, which is then negated from the original model to induce forgetting. This approach, known as Forgetting via Negation, has been shown to be effective in making LLMs unlearn harmful language generation or fail at performing specific tasks. ULD~\citep{ji2024reversing} utilizes an auxiliary LLM to achieve the unlearning objective during the decoding process of an LLM. The auxiliary LLM is trained to actively memorize the forget set while simultaneously forgetting the retain set. The unlearned LLM is generated by calculating the logit difference between the auxiliary LLM and the original model, thereby effectively achieving the unlearning objective. FILA~\citep{cha2025fila} employs LoRA adapters~\citep{lermen2024lora} to improve the computational efficiency of LLM unlearning. To achieve this, FILA identifies parameters associated with the forget set and initializes the LoRA adapters to be strongly correlated with the forget set, while the base layer is initialized to be closely related to the retain set. Subsequently, the LoRA adapters are fine-tuned using unlearning loss functions. FILA is the most related work to our study, as we also focus on achieving parameter-efficient unlearning.
\section{Preliminaries}\label{sec:3_preliminaries}

\subsection{Problem Definition}\label{subsec:3_1_problem_definition}
The goal of unlearning is to effectively eliminate the knowledge associated with a specified forget set \( \mathcal{D}_f \) in the LLM, without retraining the model from scratch. At the same time, the model is expected to preserve its performance on a retain set \( \mathcal{D}_r \), which contains knowledge that must be maintained. This objective can be formulated as an optimization problem as:
\begin{equation}
\min_{\theta} \,\, \mathbb{E}_{(x, y) \in \mathcal{D}_f} \left[ \mathcal{L}_f(y \mid x; \theta) \right] + \lambda \, \mathbb{E}_{(x, y) \in \mathcal{D}_r} \left[ \mathcal{L}_r(y \mid x; \theta) \right].
\end{equation}
In this formulation, \( \mathcal{L}_f \) denotes the loss function applied to the forget set \( \mathcal{D}_f \), encouraging the model to remove the corresponding knowledge. On the other hand, \( \mathcal{L}_r \) is the loss function applied to the retain set \( \mathcal{D}_r \), which ensures that essential knowledge is preserved. The model parameters are represented by \( \theta \), which are updated during the unlearning process. The hyperparameter \( \lambda \) controls the strength of the retention loss term, effectively regulating how strongly the model is penalized for deviating from the retain set.

\subsection{FILA: Fisher-Initialization of Low-rank Adapters}
\label{subsec:3_2_fila}

FILA achieves parameter-efficient unlearning by employing LoRA to identify parameters critical to the forget set and focuses updates on these parameters during unlearning. The overall procedure is as follows.

\textbf{Low-rank Adaptation (LoRA).}
LoRA approximates the parameter update \( \Delta W \) of a model's base weight matrix \( W \) by training an adapter composed of two low-rank matrices, \( B \) and \( A \), such that \( \Delta W = BA \). The adapter is then added to \( W \) to produce the final model. Since \( B \) and \( A \) contains far fewer parameters than \( W \), this approach enables efficient fine-tuning of LLMs with substantially fewer computational cost. 

\textbf{Forget Importance Map Extraction.}
FILA employs Fisher information (FI) to identify parameters associated with the forget set. The FI of a dataset \( \mathcal{D} \) with respect to model parameters \( \theta \) is defined as:
\begin{equation}
\mathcal{F}_{\theta}(\mathcal{D}) = \mathbb{E}_{\mathcal{D}} \left[ \left( \frac{\partial}{\partial \theta} \log p_{\theta}(\mathcal{D}) \right)^2 \right] 
\approx \frac{1}{|\mathcal{D}|} \sum_{x \in \mathcal{D}} \left( \frac{\partial}{\partial \theta} \mathcal{L}_{\text{LM}}(x; \theta) \right)^2.\label{eq:FI}
\end{equation}
\( \mathcal{L}_{\text{LM}} \) denotes the next-token prediction loss used in the pre-trained language model. The FI measures the variance of the \textbf{score function}, which is the gradient of the log-likelihood with respect to the model parameters. Intuitively, it captures how sensitively the model output changes in response to perturbations in each parameter. A higher FI value indicates that the parameter plays a more critical role in modeling the dataset \( \mathcal{D} \).

Based on this interpretation, FILA computes the ratio of FI values obtained from the forget set and retain set to determine how important each parameter is with respect to the forget set. This ratio is referred to as the \textbf{forget importance map}, denoted as \( \mathcal{M}(\mathcal{D}) \):
\begin{equation}
\mathcal{M}(\mathcal{D}) = \frac{\mathcal{F}_\theta(\mathcal{D}_f)}{\mathcal{F}_\theta(\mathcal{D}_r)} .
\end{equation}
The computed \( \mathcal{M}(\mathcal{D})\) plays a critical role in assigning weights to important parameter in LoRA-based efficient unlearning.

\textbf{LoRA Initialization with Forget Importance Map.}
FILA modifies the initialization of both the base layer and the LoRA adapter in a way that is suitable for unlearning by leveraging the forget importance map.
First, FILA formulates the following Weighted Low-Rank Approximation (WLRA) objective to obtain \( B^* \) and \( A^* \):
\begin{equation}
B^*, A^* = \arg\min_{B,\, A} \sum_{i,j} \left( \left[ \mathcal{M} \right]_{i,j} (W - BA)_{i,j} \right)^2.
\end{equation}
Since the forget importance map \( \mathcal{M} \) assigns larger weights in WLRA to parameters more relevant to the forget set, the resulting product \( B^*A^* \) captures the components of original weight matrix \( W \) that have significant influence on the forget set. Based on this, we initialize the LoRA matrices B and A with \( B^* \) and \(A^*\), respectively, so that the adapter focuses on forget-set-related parameters. FILA then obtains \( W^* \) by subtracting \( B^*A^* \) from \( W \), using it as the new base layer. Since \( B^*A^* \) concentrates information specific to the forget set, the subtraction \( W^* = W - B^*A^* \) removes forget-set-related parameters while preserving those relevant to the retain set.

Through this initialization of both the adapter and the base layer, the overall model parameters remain unchanged, as  \( W = (W - B^*A^*) + B^*A^*\). However, the information associated with the forget and retain sets becomes cleanly disentangled.

\textbf{Parameter-efficient Unlearning.} 
After initializing LoRA, FILA freezes the base layer and updates only the LoRA adapter parameters using an unlearning loss. Since parameters crucial for the forget set are allocated to the trainable adapter, while those important to the retain set remain in the frozen base layer, the model can effectively erase the undesired information while preserving essential knowledge. The final unlearned model is obtained by merging the updated adapter—now purged of forget set information—with the base layer.
\section{VILA: The Proposed Method}\label{sec:4_proposed_method}
\begin{table}[t]
\centering
\small
\label{tab:time-cost}
\begin{tabular}{lccc}
\toprule
\textbf{Method} & \textbf{Forget 1\%} & \textbf{Forget 5\%} & \textbf{Forget 10\%} \\
\midrule
\textbf{Retrain}  & 2.28 & 2.18 & 2.08 \\
\midrule
\textbf{FILA} - $\mathcal{M}(\mathcal{D})$ extraction & 0.25 & 1.21 & 9.10 \\
\textbf{FILA} - Unlearning  & 0.02 & 0.06 & 0.12 \\
\bottomrule
\end{tabular}
\caption{\textbf{Time Costs.} This table reports the GPU hours required for Retrain, Unlearn, and FILA on the TOFU benchmark using the Llama2-7B model. The Forget N\% setting indicates that N\% of the full dataset is designated as the forget set. For Retrain, the model is trained from scratch using only the retain set, which consists of the remaining (100–N)\% of the data.}\label{tab:Tab1_timecost}
\end{table}

\subsection{Corrected Parameter Importance Estimation}\label{subsec:4_1_corrected_forget_importance_map_estimation}

We argue that the forget importance map calculated by FILA is inaccurate. FILA estimates the Fisher information (FI) of each parameter with respect to the forget set and the retain set, then interprets FI as a variance measure to derive the forget importance map based on the ratio of these FIs. However, this approach overlooks a critical assumption required to interpret FI as a variance: the expectation of the score function (i.e., the gradient) must be zero~\citep{fisher1922mathematical}. This condition holds only when the distribution of the forget set matches the distribution of the entire training data. In machine unlearning tasks, however, the forget set is typically a subset of data that has been intentionally selected for removal, making its distribution inherently different from that of the entire dataset. As a result, the score function has a non-zero expectation, violating the necessary assumption. To reliably identify parameters strongly associated with the forget set, it is essential to account for distributional discrepancies that arise in unlearning scenarios.

To address this issue, we correct the FI (Equation~\ref{eq:FI}) by explicitly subtracting the squared expectation of the score function from the original formulation. This is equivalent to the variance of the parameter $\Delta W$ to the dataset $\mathcal{D}$:
\begin{equation}
\mathrm{Var}_{\mathcal{D}}[\Delta W] :=\mathbb{E}_\mathcal{D} \left[ \left( \frac{\partial}{\partial W} \log p_{W}(\mathcal{D}) \right)^2 \right] - \left( \mathbb{E}_\mathcal{D} \left[ \frac{\partial}{\partial W} \log p_{W}(\mathcal{D}) \right] \right)^2.
\end{equation}
We regard this modified quantity as an adjusted importance score for the dataset. Experimental results demonstrate that this modification significantly improves unlearning performance. While the solution is simple, identifying and correctly addressing this overlooked aspect in existing work constitutes one of the key contributions of this paper.

\subsection{Improving Efficiency via Low-rank Approximation}
\label{subsec:improved_efficiency}
Despite its intended goal, FILA is not computationally efficient. While FILA aims to perform parameter-efficient unlearning by adopting LoRA, it still requires access to the entire set of LLM parameters to compute the forget importance map. As a result, the importance map calculation remains computationally expensive.

We report the computational time required for model retraining, importance map extraction using FILA, and model unlearning in Table~\ref{tab:Tab1_timecost}, empirically demonstrating these inefficiencies. Notably, the extraction of the forget importance map, intended as a preprocessing step, incurs even greater computational cost than the unlearning process itself. This inefficiency becomes especially pronounced when the forget set constitutes approximately 10\% of the training data, where importance map computation exceeds the time required for retraining. These results suggest that FILA does not scale well with forget set size, making it suboptimal in terms of efficiency. Thus, achieving truly efficient unlearning necessitates a more efficient approach to extracting the forget importance map.

To address this issue, we propose an approach that utilizes the gradients of LoRA adapter rather than those of the entire model. First, we initialize the LoRA adapter matrices \( B \) and \( A \) independently, following a Gaussian distribution with a mean of zero. Next, we add the adapter \( BA \) to the original model parameter \( W \) and compute the gradients of \( B \) and \( A \) for a given input data \( \mathcal{D} \). Using these gradients, we calculate \( \mathrm{Var}_{\mathcal{D}}[\Delta B] \) and \( \mathrm{Var}_{\mathcal{D}}[\Delta A] \), respectively. We then multiply these two values to obtain the variance of the model parameters \( {W} \):
\begin{equation}
\mathrm{Var}_{\mathcal{D}} [\Delta W]\approx \mathrm{Var}_{\mathcal{D}}[\Delta B] \mathrm{Var}_{\mathcal{D}}[\Delta A].
\end{equation}
One critical aspect of our approach is understanding how the variance of the gradient of the model parameter can be approximated with those of LoRA adapters. To explore this, we present the following theorem:
\begin{theorem}[Variance Approximation of LoRA Parameter Updates]
Let \( \mathcal{D} \) be the input data, \( W \) be the model parameter matrix and let \( \Delta W \) denote its update. In the LoRA framework, the parameter update \( \Delta W \) is represented as the product of two low-rank matrices \( B \) and \( A \) such that:
\begin{equation}
\Delta W = BA.
\end{equation}
Assuming that both \( B \) and \( A \) are independently initialized from zero-mean Gaussian distributions, the variance of each element \( \Delta W_{ij} \) can be approximated as:
\begin{equation}
\mathrm{Var}_{\mathcal{D}} [\Delta W] \approx \mathrm{Var}_{\mathcal{D}}[\Delta B] \mathrm{Var}_{\mathcal{D}}[\Delta A].
\end{equation}
\end{theorem}

\begin{proof}
The proof is provided in the Appendix~\ref{asec:proof}.
\end{proof}
Finally, we derive the forget importance map as the element-wise ratio of the importance values calculated for the forget set and the retain set:
\begin{equation}
\mathcal{M}(\mathcal{D}) = \frac{\mathrm{Var}_{\mathcal{D}_f}[\Delta W]}{\mathrm{Var}_{\mathcal{D}_r}[\Delta W]} 
\approx 
\frac{\mathrm{Var}_{\mathcal{D}_f}[\Delta B] \mathrm{Var}_{\mathcal{D}_f}[\Delta A]}
{\mathrm{Var}_{\mathcal{D}_r}[\Delta B] \mathrm{Var}_{\mathcal{D}_r}[\Delta A]}
\end{equation}
In this way, we efficiently compute the forget importance map without directly accessing the entire parameter set of the LLM. The pseudo code of unlearning process is in Algorithm~\ref{algo:efficient_unlearning}.

\begin{algorithm}[t]
\caption{Unlearning Process}
\begin{algorithmic}[1]
\footnotesize
\STATE \textbf{Input:} Forget set \(\mathcal{D}_f\), retain set \(\mathcal{D}_r\)
\STATE \textbf{Output:} Unlearned model \( W_{\text{unlearn}} \)
\\
\STATE \textbf{Step 1: Estimate Variance-Based Importance}
\STATE \quad Initialize LoRA matrices: \( B \sim \mathcal{N}(0, \sigma^2),\ A \sim \mathcal{N}(0, \sigma^2) \)
\STATE \quad Compute gradients of \( B \) and \( A \) with respect to \( \mathcal{D}_f \) and \( \mathcal{D}_r \)
\STATE \quad Calculate variances:
\begin{equation}
\mathrm{Var}_{\mathcal{D}_f}[\Delta B], \quad \mathrm{Var}_{\mathcal{D}_f}[\Delta A], \quad
\mathrm{Var}_{\mathcal{D}_r}[\Delta B], \quad \mathrm{Var}_{\mathcal{D}_r}[\Delta A]
\end{equation}
\STATE \quad Estimate importance map:
\begin{equation}
\mathcal{M} (\mathcal{D}) = \frac{\mathrm{Var}_{\mathcal{D}_f}[\Delta B] \mathrm{Var}_{\mathcal{D}_f}[\Delta A]}{\mathrm{Var}_{\mathcal{D}_r}[\Delta B] \mathrm{Var}_{\mathcal{D}_r}[\Delta A]}
\end{equation}

\STATE \textbf{Step 2: Compute Weighted Low-Rank Approximation}
\STATE \quad Solve WLRA using \( \mathcal{M} \):
\begin{equation}
B^*, A^* = \arg\min_{B, A} \sum_{i,j} \left( \mathcal{M}_{ij} (W - BA)_{ij} \right)^2
\end{equation}
\STATE \quad Initialize LoRA adapter: \texttt{lora\_A} \(\leftarrow A^*\), \texttt{lora\_B} \(\leftarrow B^*\)
\STATE \quad Set base layer \(W^*\): \texttt{base\_layer} \(\leftarrow W - B^* A^*\)
\STATE \textbf{Step 3: Perform Unlearning}
\STATE \quad Freeze base layer \( W^*\)
\STATE \quad Optimize LoRA parameters:
\begin{equation}
B', A' = \arg\min_{B^*, A^*} \; \mathbb{E}_{(x,y) \in \mathcal{D}_f} \left[ \mathcal{L}_f(y \mid x; \theta) \right] + \lambda \, \mathbb{E}_{(x,y) \in \mathcal{D}_r} \left[ \mathcal{L}_r(y \mid x; \theta) \right]
\end{equation}

\STATE \textbf{Return:} Final unlearned model: \(W_{\text{unlearn}} = W^* + B'A'\)
\end{algorithmic}
\label{algo:efficient_unlearning}
\end{algorithm}

\section{Experiments}

\textbf{Benchmarks and compared methods.} We evaluate unlearning performance using three benchmarks: TOFU~\citep{maini2024tofu}, WMDP~\citep{li2024wmdp}, and MUSE~\citep{shi2025muse}, and primarily compare our method against FILA with three unlearning loss functions: GD~\citep{liu2022right}, NPO~\citep{zhang2024negative}, and IHL~\citep{cha2025fila}. Further details on benchmarks and compared methods are provided in Appendix~\ref{asec:benchmarks} and \ref{asec:compared_methods}, respectively.

\textbf{Implementation details.} 
All experiments are conducted using two NVIDIA A6000 GPUs with 48GB of memory. The batch size is set to 32 for TOFU and MUSE, and 4 for WMDP.
The LoRA rank is set to 8 for TOFU and WMDP, and 16 for MUSE. Weight decay is configured as 0.01 for TOFU and set to 0 for both MUSE and WMDP. We employ a linear learning rate scheduler for WMDP and TOFU, and a constant scheduler for MUSE. 

\textbf{Fair and comprehensive experimental designs.} To ensure a fair comparison, we conduct the same number of hyperparameter searches for all compared methods. Specifically, we perform random search~\citep{bergstra2012random} within a predefined hyperparameter range for each benchmark. We set the maximum unlearning epoch based on retraining cost considerations. Furthermore, to avoid evaluating models with significantly degraded utility, we select models that maintain at least 95\% of the original model utility~\citep{ilharco2023editing} while achieving the highest forgetting score. As the compared methods demonstrate comparable model utility, we report only the forgetting performance in the following tables. Additional details are provided in Appendix~\ref{asec:exp_designs}.

\begin{table}[t]
\centering
\small
\begin{tabular}{
  >{\centering\arraybackslash}m{2.3cm}|
  >{\arraybackslash}m{2.3cm}|
  >{\centering\arraybackslash}m{1.6cm}
  >{\centering\arraybackslash}m{1.6cm}
  >{\centering\arraybackslash}m{1.6cm}|
  >{\centering\arraybackslash}m{2.0cm}
}
\toprule
\textbf{Model} & \textbf{Method} & \textbf{Forget 1\%} & \textbf{Forget 5\%} & \textbf{Forget 10\%} & \textbf{AVG Gain} ($\uparrow$) \\
\midrule
\multirow{10}{*}{\centering\arraybackslash\textbf{Phi-1.5B}}
  & \textbf{Original Model}  & -4.05 & -11.92 & -15.66 & -     \\
  \cmidrule{2-6}
  & GD              & -2.52 & -11.18 & -14.43 & -     \\
  & GD + FILA       & -2.17 & -10.23 & -13.84 & 0.63  \\
  \rowcolor{gray!10}
  \cellcolor{white} & GD + Ours       & -1.54 & -9.61  & -10.80 & \textbf{2.06}  \\
  \cmidrule{2-6}
  & NPO             & -2.52 & -7.89  & -10.03 & -     \\
  & NPO + FILA      & -2.17 & -6.09  & -8.83  & 1.12  \\
  \rowcolor{gray!10}
  \cellcolor{white} & NPO + Ours      & -2.17 & -5.17  & -9.30  & \textbf{1.27}  \\
  \cmidrule{2-6}
  & IHL             & -2.52 & -10.23  & -14.13 & -     \\
  & IHL + FILA      & -2.17 & -5.40  & -1.79  & 5.84  \\
  \rowcolor{gray!10}
  \cellcolor{white} & IHL + Ours      & -1.85 & -1.17  & -0.83  & \textbf{7.68}  \\
\midrule
\multirow{10}{*}{\centering\arraybackslash\textbf{Llama2-7B}}
  & \textbf{Original Model}  & -3.30 & -15.46 & -19.31 & -     \\
  \cmidrule{2-6}
  & GD              & -3.30 & -9.92  & -16.61 & -     \\
  & GD + FILA       & -3.30 & -12.53 & -17.27 & -1.09 \\
  \rowcolor{gray!10}
  \cellcolor{white} & GD + Ours       & -2.17 & -1.40  & -1.18  & \textbf{8.36}  \\
  \cmidrule{2-6}
  & NPO             & -3.30 & -13.59 & -13.84 & -     \\
  & NPO + FILA      & -3.30 & -11.18 & -11.06 & 1.73  \\
  \rowcolor{gray!10}
  \cellcolor{white} & NPO + Ours      & -1.54 & -4.32  & -4.59  & \textbf{6.76}  \\
  \cmidrule{2-6}
  & IHL             & -3.30 & -12.53  & -7.70  & -     \\
  & IHL + FILA      & -3.30 & -0.95  & -0.47  & 6.27  \\
  \rowcolor{gray!10}
  \cellcolor{white} & IHL + Ours      & -1.27 & -0.20  & -0.40  & \textbf{7.22}  \\
\bottomrule
\end{tabular}
\begin{tabular}{
  >{\arraybackslash}m{2.3cm}|
  >{\centering\arraybackslash}m{1.3cm}
  >{\centering\arraybackslash}m{1.3cm}|
  >{\centering\arraybackslash}m{1.3cm}|
  >{\centering\arraybackslash}m{1.7cm}
  >{\centering\arraybackslash}m{1.7cm}|
  >{\centering\arraybackslash}m{1.365cm}
}
\toprule
\multirow{2}{*}{\textbf{Method}} & \multicolumn{3}{c|}{\textbf{WMDP}} & \multicolumn{3}{c}{\textbf{MUSE BOOKS}} \\ \cmidrule{2-7} 
                               & \textbf{BIO} ($\downarrow$) & \textbf{CYB} ($\downarrow$) & \textbf{AVG} ($\downarrow$) & \textbf{VerbM} ($\downarrow$) & \textbf{KnowM} ($\downarrow$) & \textbf{AVG} ($\downarrow$) \\ \midrule
\textbf{Original Model}         & 0.64              & 0.44              & 0.54               & 85.5              & 30.5              & 58.0              \\ \midrule
GD                & 0.55              & 0.44              & 0.50              & 84.7              & 17.0              & 50.9              \\ 
GD + FILA         & 0.61              & 0.44              & 0.53              & 85.4              & 17.0              & 51.2              \\ 
\cellcolor{gray!10}GD + Ours       & \cellcolor{gray!10}0.37              & \cellcolor{gray!10}0.43              & \cellcolor{gray!10}\textbf{0.40}              & \cellcolor{gray!10}66.4              & \cellcolor{gray!10}16.5              & \cellcolor{gray!10}\textbf{41.5}              \\ \midrule
NPO               & 0.57              & 0.44              & 0.51              & 14.3              & 0.3              & 7.3              \\ 
NPO + FILA        & 0.61              & 0.43              & 0.52              & 85.0              & 12.0              & 48.5              \\ 
\cellcolor{gray!10}NPO + Ours      & \cellcolor{gray!10}0.35              & \cellcolor{gray!10}0.43              & \cellcolor{gray!10}\textbf{0.39}              & \cellcolor{gray!10}2.8  & \cellcolor{gray!10}3.7 & \cellcolor{gray!10}\textbf{3.3} \\ \midrule
IHL               & 0.60              & 0.40              & 0.50              & 85.4              & 14.9              & 50.2              \\ 
IHL + FILA        & 0.61              & 0.39              & 0.50              & 85.4              & 16.7              & 51.1              \\ 
\cellcolor{gray!10}IHL + Ours      & \cellcolor{gray!10}0.48              & \cellcolor{gray!10}0.41              & \cellcolor{gray!10}\textbf{0.45}              & \cellcolor{gray!10}14.2 & \cellcolor{gray!10}1.8 & \cellcolor{gray!10}\textbf{8.0} \\ \bottomrule
\end{tabular}
\caption{\textbf{Main Comparison Results.} Top: Unlearning performance on TOFU with Phi-1.5B and Llama2-7B across varying forget ratios. AVG Gain (↑) denotes the average improvement in unlearning loss from each initialization method, measured across data splits.
Bottom: Unlearning performance on WMDP and MUSE Books. AVG is the mean of the two forget metrics per benchmark. Lower scores indicate better forgetting performance.
Retain performance is omitted, as it is constrained to remain above 95\% by our evaluation protocol.}
\label{tab:Table2_main_results}
\end{table}

\subsection{Comparison with FILA}

This section presents the results on the TOFU, WMDP, and MUSE benchmarks. Note that rows labeled as GD, NPO, or IHL refer to standard parameter-efficient unlearning baselines that utilize a conventional LoRA adapter without applying either FILA or our method.

\textbf{Main Results.} 
In Table~\ref{tab:Table2_main_results}, we observe that both FILA and our proposed method significantly improve forgetting performance compared to applying unlearning loss functions with standard LoRA alone. This highlights the effectiveness of explicitly separating the parameters associated with the forget set, rather than relying on full-parameter updates.

While both approaches benefit from this separation, our method consistently achieves superior forgetting quality across all benchmarks and loss functions. FILA, in contrast, performs well primarily when paired with IHL—the loss function introduced alongside it by \citet{cha2025fila}—but offers limited gains with GD or NPO, and in some cases even degrades performance. For example, on Llama2-7B with GD, applying FILA results in lower forgetting quality when unlearning 5\% or 10\% of the TOFU dataset. This aligns with the analysis in the original submission on OpenReview~\citep{cha2025fila}, which points to FILA’s limited generalizability beyond its tailored loss. In contrast, our method not only outperforms FILA when paired with IHL but also maintains strong performance across diverse unlearning losses, demonstrating its broader applicability.

Our method yields substantial improvements when applied to GD, where the forget loss is implemented via GA. Although GA often leads to instability and utility degradation~\citep{zhang2024negative, cha2025fila}, our method effectively mitigates these issues and achieves strong forgetting performance. This result is also significant in MUSE, where the forget set size exceeds the retain set size, making it particularly difficult to preserve model utility. Despite this difficulty, our method substantially outperforms existing baselines, delivering up to a 9.4\%p increase in performance (50.9\% $\rightarrow$ 41.5\%) when combined with GD.

\begin{table}[t]
\centering
\small
\label{tab:time_storage_comparison}
\begin{tabular}{l|ccc|c}
\toprule
\textbf{Method} & \multicolumn{3}{c|}{\textbf{Time (GPU hours)}} & \textbf{Storage for $\mathcal{M}(\mathcal{D})$} \\
\cmidrule(lr){2-4}
& Forget 1\% & Forget 5\% & Forget 10\% & \\
\midrule
Retrain & 2.28 & 2.18 & 2.08 & -- \\
\midrule
FILA   & 0.27 & 1.27 & 9.22 & 25G \\
Ours   & \textbf{0.04} & \textbf{0.18} & \textbf{0.36} & \textbf{0.3G} \\
\bottomrule
\end{tabular}
\caption{\textbf{Comparison of time and storage cost across different methods.} We report GPU hours required for unlearning under varying forget set sizes (1\%, 5\%, 10\%) on the TOFU benchmark using the Llama2-7B model. Storage for \(\mathcal{M}(\mathcal{D})\) 
denotes the additional space required to store forget information map used during unlearning.}
\label{tab:Table5_totalcost}
\end{table}

\textbf{TOFU Results.} The Forget 1\% setting in TOFU is particularly challenging compared to other configurations. In this setting, only 1\% of the entire dataset is designated as the forget set, meaning the model must remove knowledge from a very small portion of the data. This also implies that only a very small number of parameters are associated with the forget set, which makes it challenging to precisely identify and update the relevant subset. Especially with Llama2-7B, applying unlearning via GD, IHL, or NPO loss alone produces forget quality scores nearly indistinguishable from those of the original model, suggesting that information related to the forget set remains. Even when applying FILA, a slight improvement in forget quality is observed with the Phi-1.5B model; however, for Llama2-7B, the forget quality remains at the same level as the target model. In contrast, our method achieves high forgetting quality while maintaining model utility, even when the forget set size is very small. We attribute this to the method’s capacity to pinpoint parameters most relevant to the forget set.

\textbf{WMDP and MUSE Results.} In the WMDP experiments, our method continues to demonstrate strong unlearning performance across all unlearning loss functions, while preserving accuracy on MMLU. In contrast, FILA consistently underperforms compared to the baselines on this benchmark, highlighting the limitations of its biased forget importance map estimation strategy. On the MUSE benchmark, models trained with FILA struggle to maintain retain performance above 95\% of the original model. When hyperparameters are chosen to effectively reduce performance on the forget set, performance on the retain set also drops noticeably. Conversely, when the retain performance is preserved, the level of unlearning is limited. This behavior indicates that FILA fails to achieve effective unlearning. By contrast, our proposed method achieves consistent gains in forgetting performance across all loss functions, while sufficiently preserving model utility. These results validate the effectiveness of our refined approach to estimating forget importance.

\subsection{Efficiency Analysis} 

Table~\ref{tab:Table5_totalcost} compares the time and storage costs for unlearning on the TOFU benchmark using Llama2-7B as the backbone model. FILA incurs substantial time costs due to the need to compute gradients over the entire model to extract the forget importance map. This becomes increasingly inefficient as the forget set size grows. For instance, FILA takes 0.27 GPU hours for the 1\% setting, but this rises sharply to 9.22 GPU hours for the 10\% setting. Notably, this exceeds the retraining time of approximately 2.08 GPU hours, underscoring FILA’s limitations in large-scale unlearning scenarios. In contrast, our method requires only 0.04 GPU hours for the 1\% setting and 0.36 GPU hours for the 10\% setting, while achieving comparable unlearning quality at a fraction of the computational cost.  

A key difference between the two methods lies in storage efficiency. Both approaches require storing importance maps for the forget and retain sets. FILA computes these maps across all model parameters, resulting in a total storage requirement of approximately 25 GB—about twice the size of the original model. In contrast, our method uses LoRA parameters instead of the full parameter set, requiring only 0.3 GB of additional storage.

\subsection{Ablation Study on FI Correction and LoRA Approximation}

To investigate how each of our two key contributions individually impacts the overall unlearning performance, we conduct ablation experiments clearly isolating the following two components: (i) \textbf{FI Correction (Eq.~5)}, which involves correcting the Fisher Information estimation, and (ii) \textbf{LoRA Approximation (Eq.~9)}, which approximates the full-model Fisher Information using LoRA adapter parameters to enhance computational efficiency.

\begin{table}[t]
\centering
\small
\begin{tabular}{l|l|ccc|c}
\toprule
Loss & Method & Forget 1\% & Forget 5\% & Forget 10\% & AVG Gain ($\uparrow$) \\
\midrule
\multirow{4}{*}{GD} 
& FILA (Baseline)       & -2.17 & -10.23 & -13.84 & —    \\
& w/ FI Correction      & {-1.27} & {-9.61}  & {-9.54}  & \textbf{1.94} \\
& w/ LoRA Approximation & -2.17 & {-9.61}  & -13.54 & 0.30 \\
\rowcolor{gray!10}
  \cellcolor{white} & w/ Both (VILA)        & -1.54 & {-9.61}  & -10.80 & 1.43 \\
\midrule
\multirow{4}{*}{NPO} 
& FILA (Baseline)       & -2.17 & -6.09  & -8.83  & —    \\
& w/ FI Correction      & {-1.85} & {-6.34}  & {-5.85}  & \textbf{1.02} \\
& w/ LoRA Approximation & -2.17 & -6.58  & -9.30  & -0.32 \\
\rowcolor{gray!10}
  \cellcolor{white} & w/ Both (VILA)        & -2.17 & -5.17  & -9.30  & 0.15 \\
\midrule
\multirow{4}{*}{IHL} 
& FILA (Baseline)       & -2.17 & -5.40  & -1.79  & —    \\
& w/ FI Correction      & {-1.54} & {-0.85}  & -0.47  & \textbf{2.17} \\
& w/ LoRA Approximation & -2.17 & -4.53  & {-0.19}  & 0.82 \\
\rowcolor{gray!10}
  \cellcolor{white} & w/ Both (VILA)        & -1.85 & -1.17  & -0.83  & 1.84 \\
\bottomrule
\end{tabular}
\caption{Ablation results isolating the impact of FI Correction and LoRA Approximation on unlearning performance. Results show Forget Quality scores (lower is better). AVG Gain denotes improvement over the FILA baseline.}
\label{tab:ablation_fi_lora}
\end{table}

Table~\ref{tab:ablation_fi_lora} summarizes detailed ablation results conducted using the Phi-1.5B model, across three unlearning loss functions (GD, NPO, IHL). From these ablation experiments, we observe the following. FI Correction alone consistently yields the highest unlearning performance across most settings, confirming the significant effectiveness of correcting Fisher Information estimation. In contrast, LoRA Approximation alone primarily enhances computational efficiency but achieves limited or no performance improvement compared to the FILA baseline. When the two components are combined—as in \textbf{VILA}—the resulting method achieves performance close to that of FI Correction alone, while significantly reducing computational overhead. This demonstrates that VILA effectively balances strong unlearning performance with computational efficiency.

\subsection{Validity of the Expectation as an Importance Score}\label{subsec:6_2_validity_of_the_expectation}
When the distribution of the forget set ($\mathcal{D}_f$) significantly differs from that of the entire dataset ($\mathcal{D}$), the expectation of the score function—specifically, the gradient of the log-likelihood with respect to the parameters—becomes non-zero. This observation raises an important question regarding the validity of using the expectation itself directly as an importance score. To empirically investigate this question, we consider two alternative formulations for the importance score, both based on the score function $\nabla_W \log p_W(\mathcal{D})$:
\begin{equation}
\mathcal{M}_{\text{ExpILA}} 
:= \frac{
    \left| \mathbb{E}_{\mathcal{D}_f} \left[ \dfrac{\partial}{\partial W} \log p_W(\mathcal{D}_f) \right] \right|
}{
    \left| \mathbb{E}_{\mathcal{D}_r} \left[ \dfrac{\partial}{\partial W} \log p_W(\mathcal{D}_r) \right] \right|
}, \quad
\mathcal{M}_{\text{AbsILA}} 
:= \frac{
    \mathbb{E}_{\mathcal{D}_f} \left[ \left| \dfrac{\partial}{\partial W} \log p_W(\mathcal{D}_f) \right| \right]
}{
    \mathbb{E}_{\mathcal{D}_r} \left[ \left| \dfrac{\partial}{\partial W} \log p_W(\mathcal{D}_r) \right| \right]
}
\end{equation}

Here, $\mathcal{M}_{\textsc{ExpILA}}$ computes the magnitude of the expected score function, whereas $ \mathcal{M}_{\textsc{AbsILA}}$ computes the expected magnitude of the score function. Intuitively, while $\textsc{ExpILA}$ focuses on the norm of the average gradient, $\textsc{AbsILA}$ accounts for the average sensitivity across data points, regardless of gradient direction cancellation.

\begin{table}[t]
\centering
\small
\begin{tabular}{l|l|ccc|c}
\toprule
Loss & Method & Forget 1\% & Forget 5\% & Forget 10\% & AVG Gain ($\uparrow$) \\
\midrule
\multirow{4}{*}{GD} 
  & FILA     & -2.17 & -10.23 & -13.84 & -- \\
  & ExpILA   & -1.27 & -10.54 & -10.54 & 1.93 \\
  & AbsILA   & -1.54 & -10.23 & -10.29 & 2.03 \\
\rowcolor{gray!10}
  \cellcolor{white} & VILA & -1.54 & {-9.61} & {-10.80} & \textbf{2.06} \\
\midrule
\multirow{4}{*}{NPO} 
  & FILA     & -2.17 & -6.09 & -8.83 & -- \\
  & ExpILA   & -1.85 & -5.62 & -10.54 & 2.74 \\
  & AbsILA   & -1.85 & -5.86 & -9.06 & 1.23 \\
\rowcolor{gray!10}
  \cellcolor{white} & VILA & {-1.85} & {-1.17} & {-0.83} & \textbf{7.68} \\
\midrule
\multirow{4}{*}{IHL} 
  & FILA     & -2.17 & -5.40 & -1.79 & -- \\
  & ExpILA   & -0.39 & -3.37 & -6.05 & 2.43 \\
  & AbsILA   & -1.27 & -3.55 & -5.11 & 5.65 \\
\rowcolor{gray!10}
  \cellcolor{white} & VILA & {-1.85} & {-1.17} & {-0.83} & \textbf{7.68} \\
\bottomrule
\end{tabular}
\caption{\textbf{Validity of the Expectation as an Importance Score on Phi-1.5B}. AVG Gain (↑) denotes average improvement in unlearning loss across splits.}
\label{tab:phi-full-results-remove-utility}
\end{table}

Experimental results in Table~\ref{tab:phi-full-results-remove-utility} summarize the trends discussed above on Phi-1.5B. From these results, we observe several important findings. Both \textsc{ExpILA} and \textsc{AbsILA} generally perform comparably to, or slightly better than, \textsc{FILA} when the forget set is small (e.g., Forget 1\%). However, as the forget set grows (e.g., Forget 5\% and 10\%), \textbf{VILA} consistently outperforms both alternatives. These results indicate that while \textsc{ExpILA} and \textsc{AbsILA} may be reasonable under limited forgetting scenarios, correcting the Fisher Information—as in \textbf{VILA}—is essential for effective unlearning in realistic scenarios where the forget set significantly diverges from the full data distribution. Similar trends are observed for Llama2-7B (Appendix~\ref{appendix:A_3_validity_of}).
\section{Conclusion}
In this paper, we introduce VILA, a scalable and efficient unlearning technique for large language models (LLMs) that addresses the limitations of FILA by enhancing importance estimation and reducing computational overhead. Our approach refines the Fisher information extraction process and implements parameter selection related to the forget set by utilizing only the gradients of the LoRA adapter, significantly lowering time and memory costs. Extensive experiments on the TOFU, WMDP, and MUSE benchmarks demonstrate that VILA consistently outperforms existing approaches in unlearning performance while preserving model utility. Additionally, VILA demonstrates robust compatibility with a wide range of unlearning loss functions, highlighting its versatility.

\section*{Acknowledgements}
This work was supported by the National Research Foundation of Korea (NRF) grant funded by the Korean government (MSIT) (RS-2024-00350430). 
We sincerely thank the researchers at Sogang University's VRL Lab—especially Minyoung Lee, Dongjun Hwang, Yeji Park, and Beomyun Kwon—for their valuable feedback and support. We are also grateful to Seong Joon Oh from the STAI Group at the University of T\"ubingen for his helpful discussions throughout the project. Finally, we would like to thank Sungmin Cha at New York University for his insightful feedback and suggestions.

\bibliography{colm2025_conference}
\bibliographystyle{colm2025_conference}

\appendix
\newpage

\section{Proof of Theorem 1}
\label{asec:proof}

\begin{proof}

\textbf{Step 1 (Parameterization and Approximation).}

We start from the LoRA parameter update:
\[
\Delta W = BA = (B_0 + \Delta B)(A_0 + \Delta A),
\]
where \(B_0, A_0\) denote the initial low-rank parameter matrices, and \(\Delta B, \Delta A\) are their updates learned during training. Given the negligible magnitude of the initial matrices \(B_0, A_0\) (assumption \textbf{A.1}), we simplify the update as:
\[
\Delta W \approx \Delta B \Delta A.
\]

Thus, each element of the parameter update \(\Delta W\) can be approximated as:
\[
\Delta W_{ij} \approx \sum_{k=1}^{r}\Delta B_{ik}\Delta A_{kj}.
\]

\medskip
\textbf{Step 2 (Variance Expansion).}

Applying the standard property of variance to the sum of random variables, we have:
\[
\mathrm{Var}_{\mathcal{D}}\left(\sum_{k=1}^{r}\Delta B_{ik}\Delta A_{kj}\right)
=\sum_{k=1}^{r}\mathrm{Var}_{\mathcal{D}}[\Delta B_{ik}\Delta A_{kj}]
+\sum_{k\neq k'}\mathrm{Cov}_{\mathcal{D}}(\Delta B_{ik}\Delta A_{kj},\Delta B_{ik'}\Delta A_{k'j}).
\]

\medskip
\textbf{Step 3 (Covariance Terms are Zero).}

We now show that all off-diagonal covariance terms vanish. Consider an arbitrary off-diagonal term (\(k \neq k'\)):
\[
\mathrm{Cov}_{\mathcal{D}}(\Delta B_{ik}\Delta A_{kj}, \Delta B_{ik'}\Delta A_{k'j})
= E_{\mathcal{D}}[\Delta B_{ik}\Delta A_{kj}\Delta B_{ik'}\Delta A_{k'j}] - E_{\mathcal{D}}[\Delta B_{ik}\Delta A_{kj}]E_{\mathcal{D}}[\Delta B_{ik'}\Delta A_{k'j}].
\]

By the independence of matrices \(\Delta B\) and \(\Delta A\) (assumption \textbf{A.2}) and independence among distinct elements within each matrix (assumption \textbf{A.3}), each expectation factorizes exactly, yielding:
\[
\mathrm{Cov}_{\mathcal{D}}(\Delta B_{ik}\Delta A_{kj}, \Delta B_{ik'}\Delta A_{k'j}) = 0.
\]

Thus, all off-diagonal covariance terms vanish, simplifying the expression to:
\[
\mathrm{Var}_{\mathcal{D}}[\Delta W_{ij}] = \sum_{k=1}^{r}\mathrm{Var}_{\mathcal{D}}[\Delta B_{ik}\Delta A_{kj}].
\]

\medskip
\textbf{Step 4 (Variance Factorization and Practical Approximation).}

Since the terms \(\Delta B_{ik}\) and \(\Delta A_{kj}\) are independent (assumption \textbf{A.2}), we approximate:
\[
\mathrm{Var}_{\mathcal{D}}[\Delta B_{ik}\Delta A_{kj}]
= E_{\mathcal{D}}[\Delta B_{ik}^2]E_{\mathcal{D}}[\Delta A_{kj}^2]
- (E_{\mathcal{D}}[\Delta B_{ik}]E_{\mathcal{D}}[\Delta A_{kj}])^2.
\]

Empirically, the squared expectations of the parameter updates are negligible compared to their variances (assumption \textbf{A.4}). Hence, the above expression simplifies into:

\[
\mathrm{Var}_{\mathcal{D}}[\Delta B_{ik}\Delta A_{kj}] \approx \mathrm{Var}_{\mathcal{D}}[\Delta B_{ik}]\mathrm{Var}_{\mathcal{D}}[\Delta A_{kj}].
\]

Consequently, the final simplified variance approximation is:
\[
\mathrm{Var}_{\mathcal{D}}[\Delta W_{ij}] \approx \sum_{k=1}^{r}\mathrm{Var}_{\mathcal{D}}[\Delta B_{ik}]\mathrm{Var}_{\mathcal{D}}[\Delta A_{kj}].
\]
\end{proof}

\newcommand{\todo}[1]{\textcolor{red}{#1}}

\subsection*{A.1 Negligible magnitude of initial matrices $B_0, A_0$}

\begin{table}[ht]
\centering
\begin{tabular}{
  >{\centering\arraybackslash}m{2.3cm}|
  >{\centering\arraybackslash}m{2.3cm}|
  >{\centering\arraybackslash}m{2.3cm}}
\toprule
\textbf{Term} & \textbf{Forget Set} & \textbf{Retain Set} \\
\midrule
$B_0A_0$                  & 0.00568  & 0.00568 \\
$\Delta B A_0$           & 39.75    & 37.0     \\
$B_0\Delta A$            & 40.5     & 38.5     \\

\rowcolor{gray!10} $\Delta B\Delta A$       & 688128.0 & 585728.0 \\
\bottomrule
\end{tabular}
\caption{\textbf{Average norm values of each term across all parameters.} The gradients are computed separately for the forget and retain sets, resulting in distinct values for gradient-dependent terms. 
Note that $B_0A_0$ does not depend on any gradients but is determined solely by the initialization, and thus yields the same value for both sets.}
\label{Table7:term_norms}
\end{table}

Our variance approximation assumes that the initial parameter matrices $B_0$ and $A_0$ in the LoRA framework have negligible magnitude, allowing their direct contributions to the parameter updates to be disregarded in our derivations.

Specifically, matrices $B_0$ and $A_0$ are independently initialized from Gaussian distributions with zero mean and very small variance (typically on the order of $10^{-4}$ to $10^{-6}$). Due to this initialization scheme, the initial magnitudes of these matrices are sufficiently small that their direct contribution terms, such as $B_0 A_0$, $B_0\Delta A$, and $\Delta B A_0$, become negligible compared to the dominant update term $\Delta B\Delta A$.

In Table~\ref{Table7:term_norms}, we empirically confirm that the magnitude of initial terms is significantly smaller than that of the learned update term. Specifically, averaged across all parameters, the norm of the dominant update term $\Delta B\Delta A$ reaches values as high as $688,128$ on forget set and $585,728$ on retain set, while the norms of other terms such as $B_0A_0$, $B_0\Delta A$, and $\Delta B A_0$ are several orders of magnitude smaller. These observations justify treating the initial-related terms as negligible when approximating the gradient dynamics.

\subsection*{A.2 Independence between update matrices $\Delta B$ and $\Delta A$}

We assume that the parameter update matrices $\Delta B$ and $\Delta A$ are statistically independent. The independence assumption can be justified if the following two conditions are met: (1) the entries of $\Delta B$ and $\Delta A$ follow Gaussian distributions, and (2) the covariance between elements of $\Delta B$ and $\Delta A$ is zero,
i.e.,
\[
\mathrm{Cov}_{\mathcal{D}}(\Delta B_{ik}, \Delta A_{kj}) = 0\quad\text{for all } i,j,k.
\]

\begin{figure}[htbp]
    \centering
    \begin{subfigure}[b]{0.48\linewidth}
        \centering
        \includegraphics[width=\linewidth]{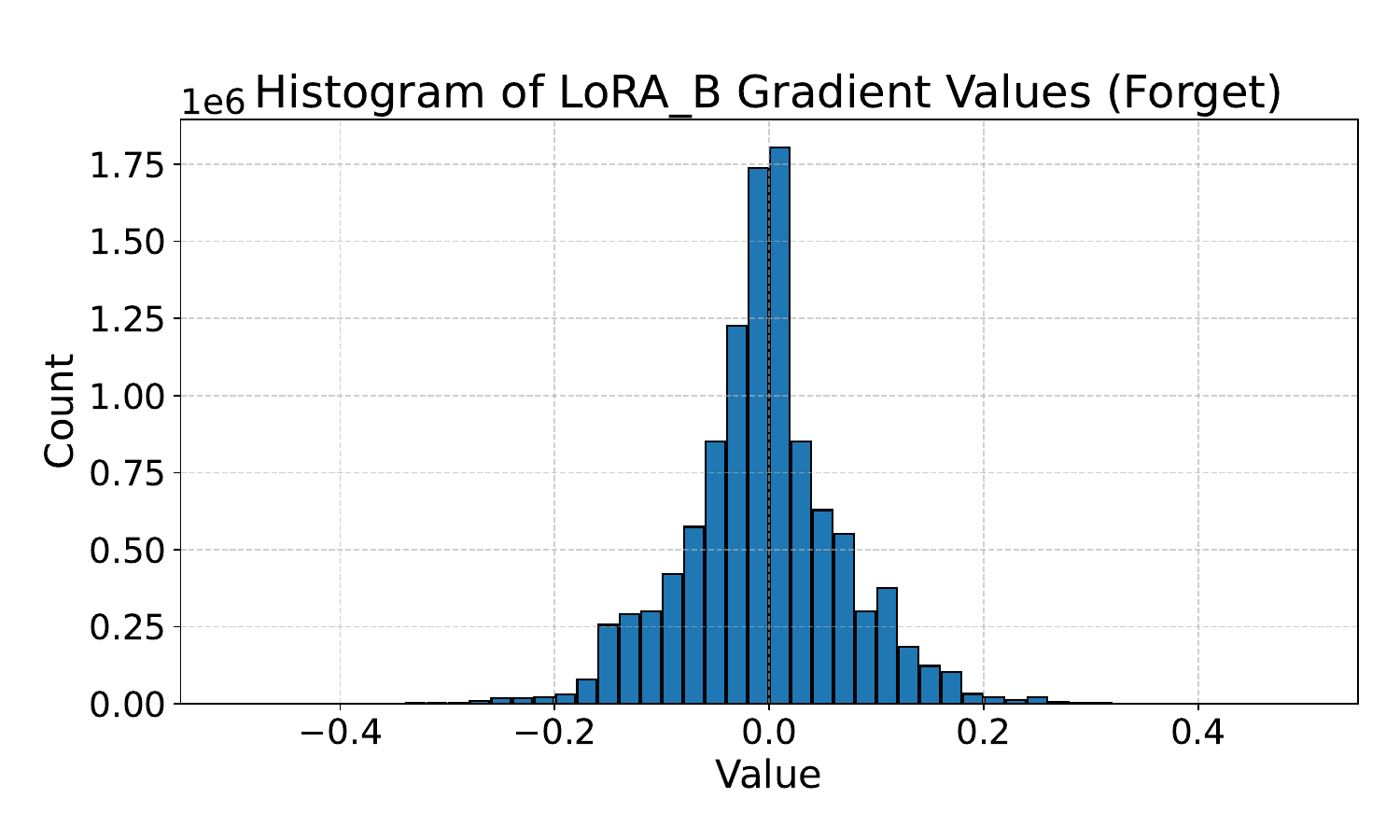}
        \caption{\(\Delta B\) on Forget set}
        \label{fig:forget_b}
    \end{subfigure}
    \hfill
    \begin{subfigure}[b]{0.48\linewidth}
        \centering
        \includegraphics[width=\linewidth]{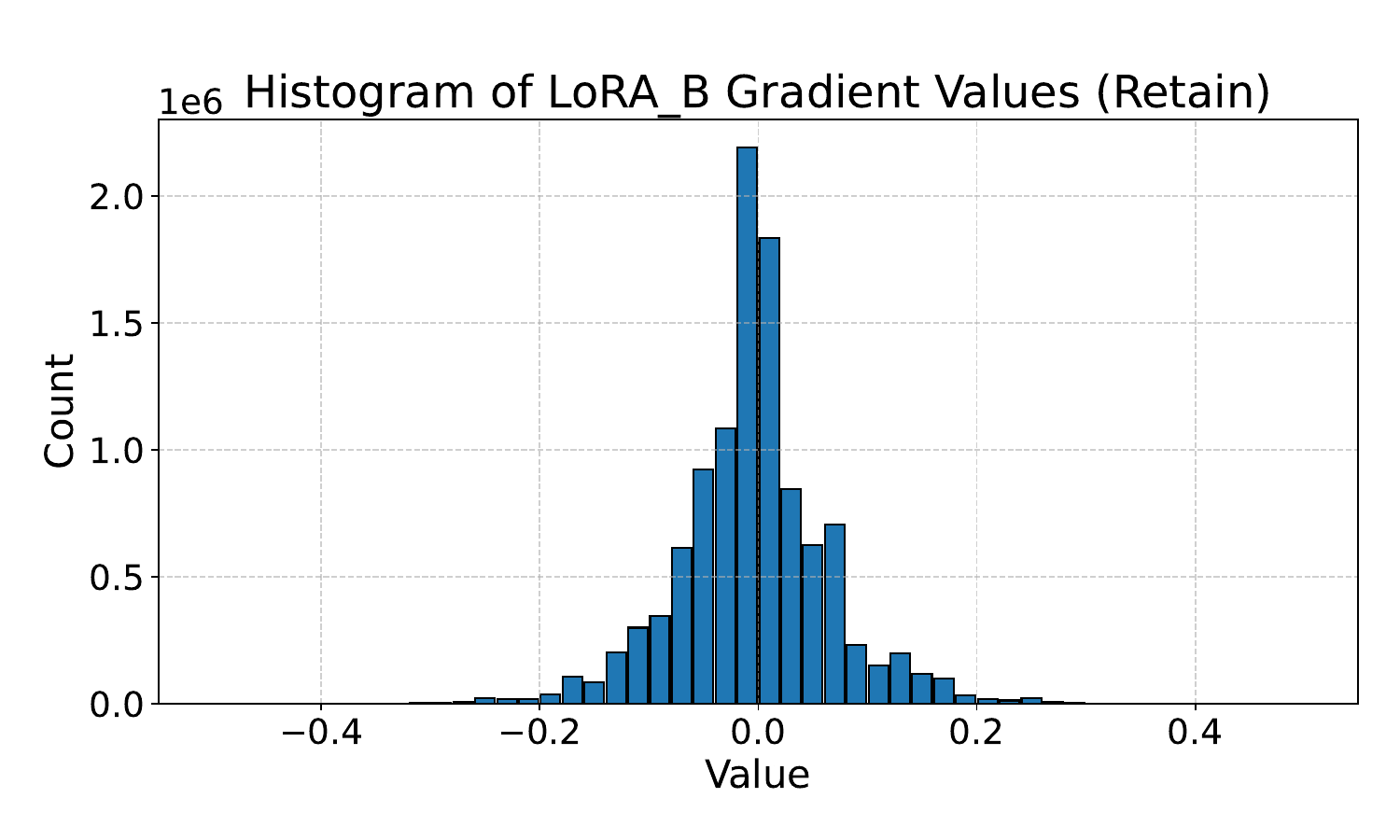}
        \caption{\(\Delta B\) on Retain set}
        \label{fig:retain_b}
    \end{subfigure}

    \vspace{0.5em} 

    \begin{subfigure}[b]{0.48\linewidth}
        \centering
        \includegraphics[width=\linewidth]{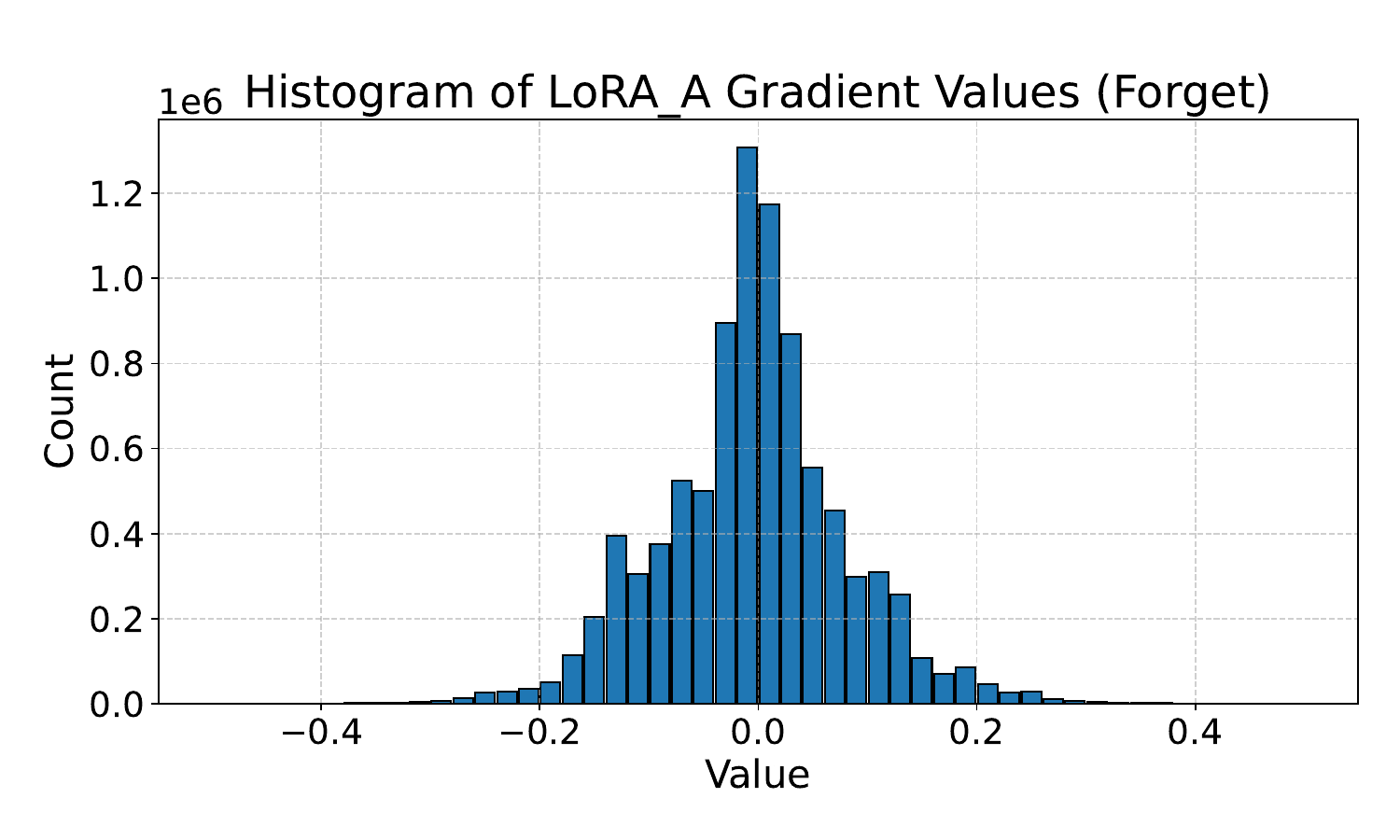}
        \caption{\(\Delta A\) on Forget set}
        \label{fig:forget_a}
    \end{subfigure}
    \hfill
    \begin{subfigure}[b]{0.48\linewidth}
        \centering
        \includegraphics[width=\linewidth]{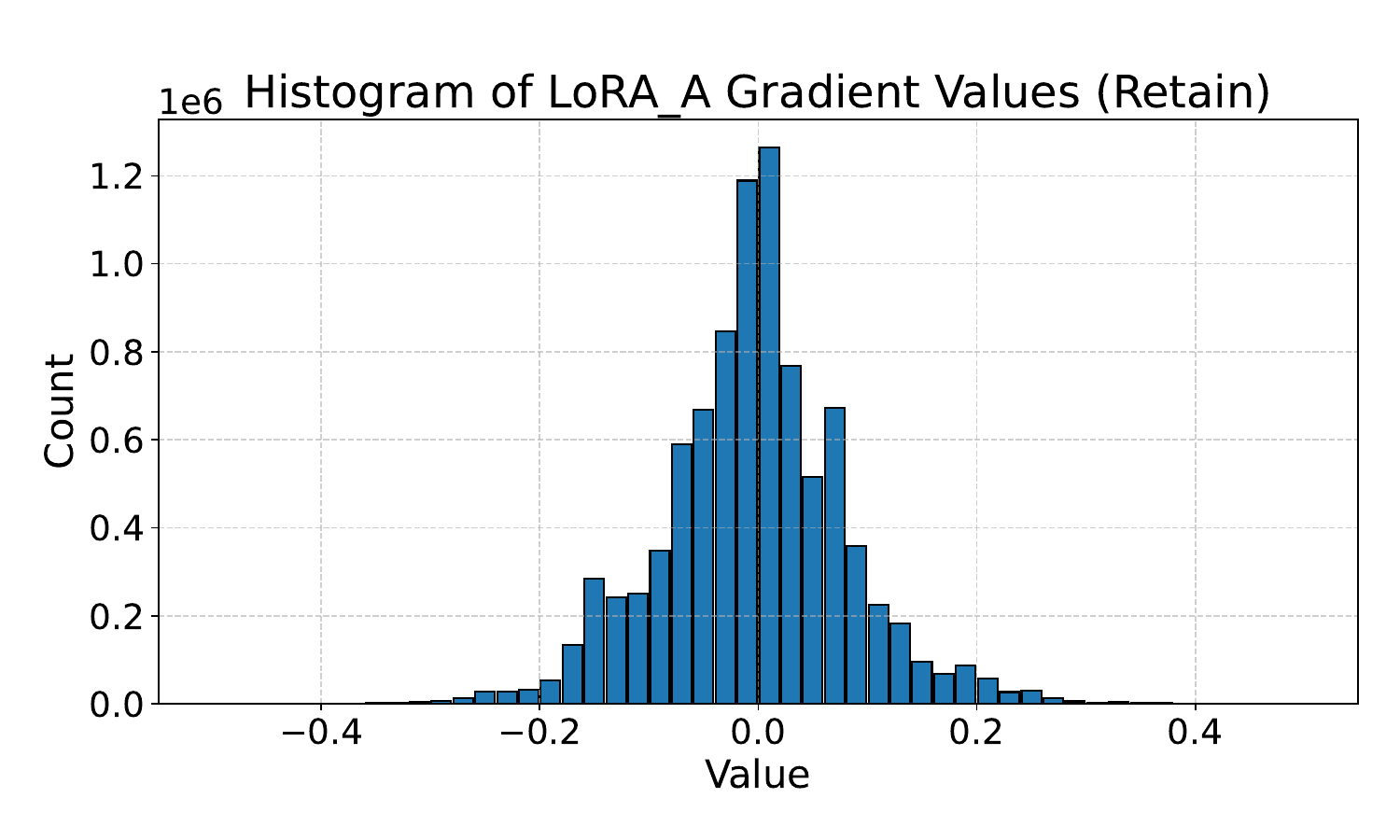}
        \caption{\(\Delta A\) on Retain set}
        \label{fig:retain_a}
    \end{subfigure}

    \caption{\textbf{Histograms of individual entries in $\Delta A$ and $\Delta B$.} }

    \label{fig:Figure6_grad_hist_all}
\end{figure}

We first examine the Gaussian assumption. This condition naturally arises from the initialization strategy used in the LoRA framework, particularly in the context of Fisher Information (FI) computation. When computing FI, the model parameters are kept fixed, and gradients are repeatedly evaluated at the same point in parameter space. To allow gradients to flow through the LoRA modules during this process, we initialize the low-rank matrices $B_0$ and $A_0$ with small Gaussian noise rather than zeros (see Appendix~\ref{sec:AppD_Sigma Ablation} for details). These initializations are independently drawn from zero-mean Gaussian distributions, which induces Gaussian in the gradient signals that propagate through $\Delta B$ and $\Delta A$. As illustrated in Figure~\ref{fig:Figure6_grad_hist_all}, the empirical distribution of gradient values indeed closely follows a Gaussian shape, confirming the plausibility of this assumption.

\begin{figure}[htbp]
    \centering
    \begin{subfigure}[b]{0.48\linewidth}
        \centering
        \includegraphics[width=\linewidth]{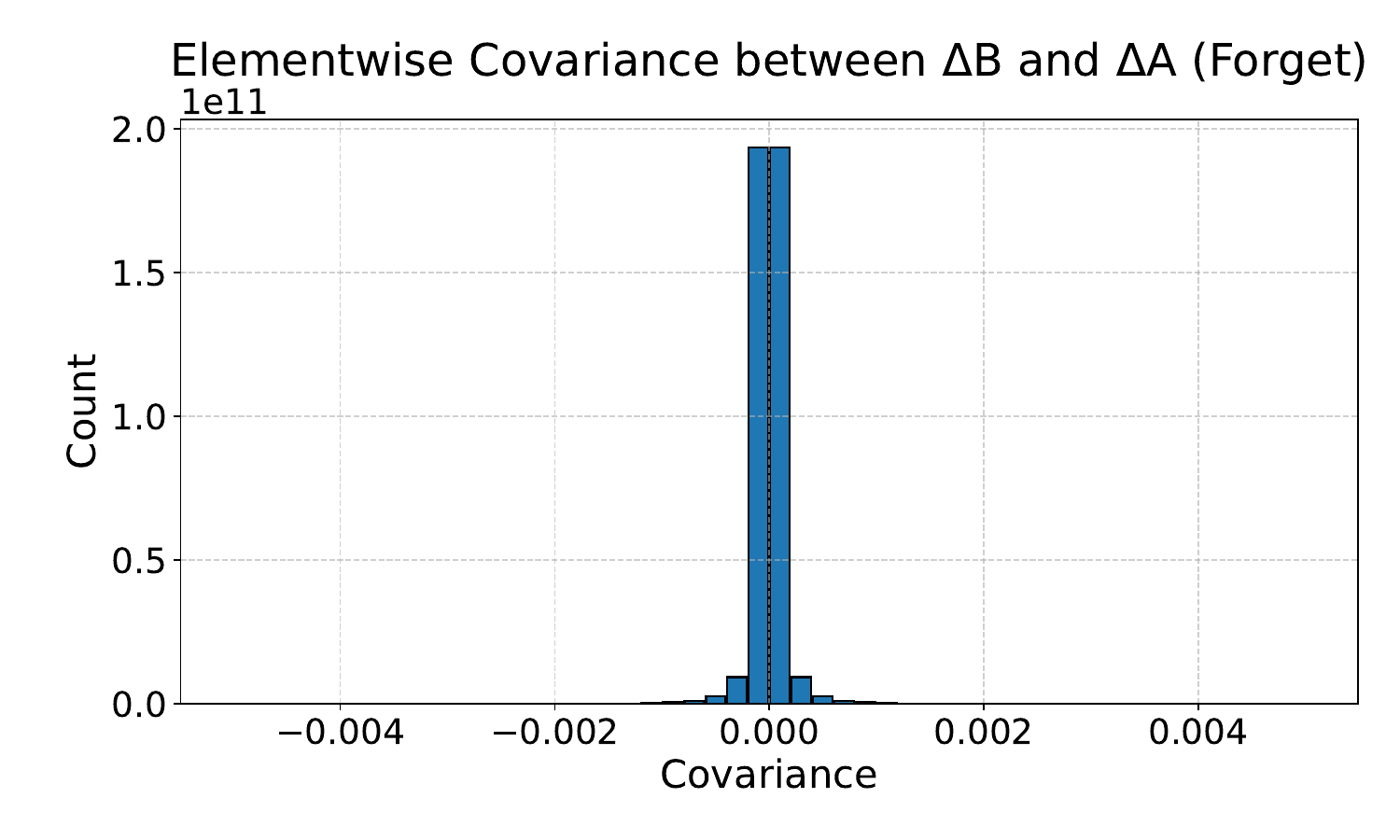}
        \caption{\(\Delta B\) and \(\Delta A\) on Forget set}
        \label{fig:left}
    \end{subfigure}
    \hfill
    \begin{subfigure}[b]{0.48\linewidth}
        \centering
        \includegraphics[width=\linewidth]{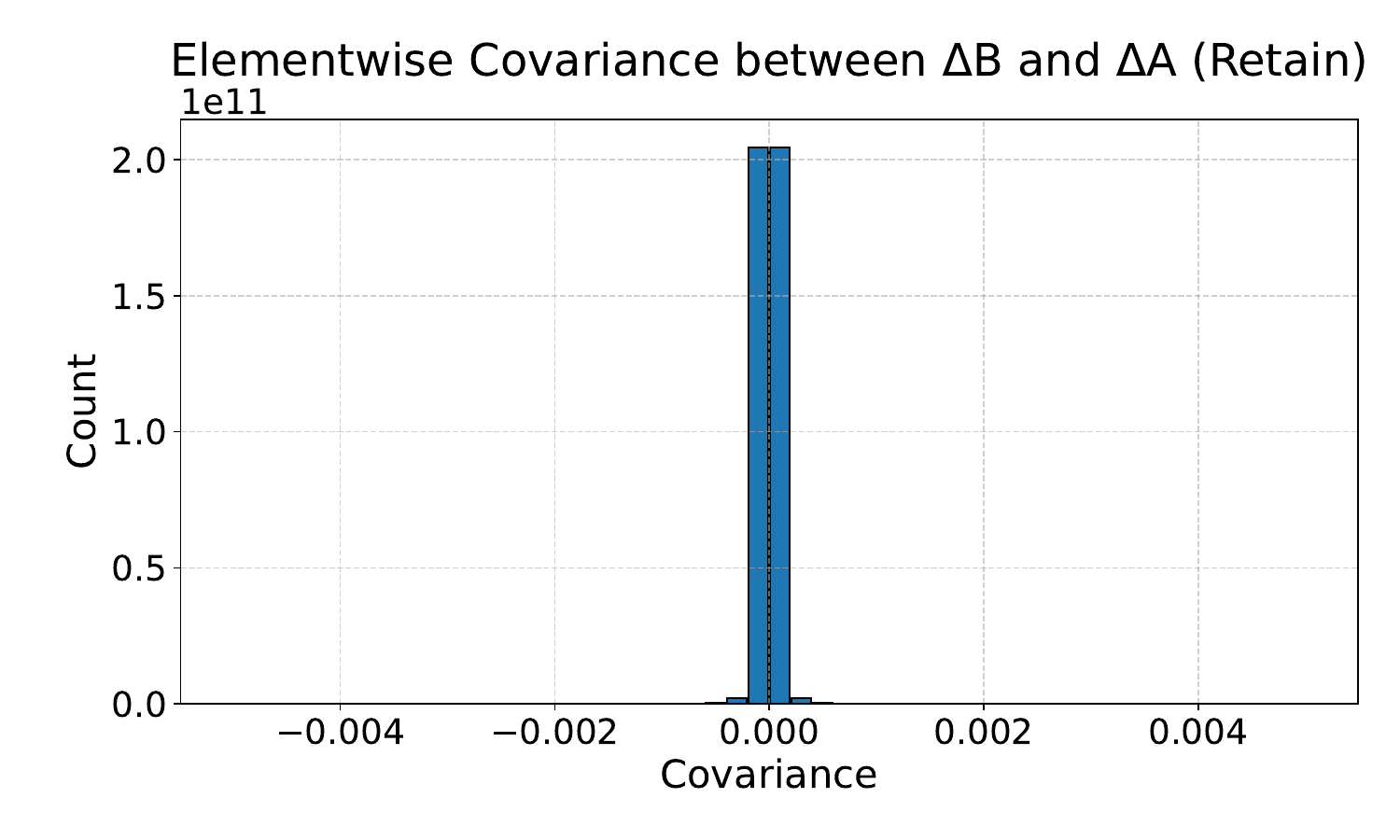}
        \caption{\(\Delta B\) and \(\Delta A\) on Retain set}
        \label{fig:right}
    \end{subfigure}
    \caption{\textbf{Element-wise covariance between $\Delta B$ and $\Delta A$.} 
We compute the covariance between all pairwise combinations of elements from the $\Delta B$ and $\Delta A$ gradient matrices, separately over the forget and retain sets. 
Each computed covariance value is counted to construct a histogram, allowing us to visualize the overall distribution of cross-matrix interactions.}

\label{fig:Figure4_covariance_histogram}
\end{figure}

Next, we assess the second requirement—vanishing covariance between elements of $\Delta B$ and $\Delta A$. To this end, we sample 500 gradient instances by repeatedly drawing different mini-batch combinations, separately for the forget and retain sets, and computing the corresponding LoRA gradients. We then calculate the element-wise covariance between $\Delta B$ and $\Delta A$ across these samples. The results, shown in Figure~\ref{fig:Figure4_covariance_histogram}, reveal that the vast majority of covariance values are sharply concentrated around zero in both cases. This confirms that any statistical dependencies between the two matrices are negligible.

Taken together, these empirical validations support the assumption that $\Delta B$ and $\Delta A$ are approximately independent. This independence greatly simplifies our variance-based theoretical derivations and is well-justified experimentally.

\subsection*{A.3 Practical independence among distinct elements within each update matrix}

In our variance approximation, we assume that distinct elements within each update matrix ($\Delta B$ and $\Delta A$) are statistically independent. This assumption is crucial for simplifying higher-order expectations. For instance, the fourth-order moment
\[
\mathbb{E}_{\mathcal{D}}[\Delta B_{ik} \Delta A_{kj} \Delta B_{ik'} \Delta A_{k'j}]
\]
is assumed to factorize as
\[
\mathbb{E}_{\mathcal{D}}[\Delta B_{ik}] \cdot \mathbb{E}_{\mathcal{D}}[\Delta A_{kj}] \cdot \mathbb{E}_{\mathcal{D}}[\Delta B_{ik'}] \cdot \mathbb{E}_{\mathcal{D}}[\Delta A_{k'j}].
\]

While Section~A.2 has already justified the independence across matrices (i.e., between $\Delta B$ and $\Delta A$), the above factorization further requires that elements within each matrix also be statistically independent. To support this intra-matrix independence, two conditions must be satisfied: (1) the elements of each matrix follow Gaussian distributions, and (2) the pairwise covariances between distinct elements within the same matrix are negligible. The first condition has already been empirically confirmed in Figure~\ref{fig:Figure6_grad_hist_all}, where we show that the elements of $\Delta B$ and $\Delta A$ follow approximately zero-mean Gaussian distributions under our sampling procedure.

To validate the second condition, we compute the element-wise covariance between all pairs of distinct entries within each matrix (i.e., off-diagonal pairs). We sample 500 gradient instances by repeatedly drawing different mini-batch combinations, separately for the forget and retain sets, and compute the corresponding LoRA gradients.

\begin{figure}[htbp]
    \centering
    \begin{subfigure}[b]{0.48\linewidth}
        \centering
        \includegraphics[width=\linewidth]{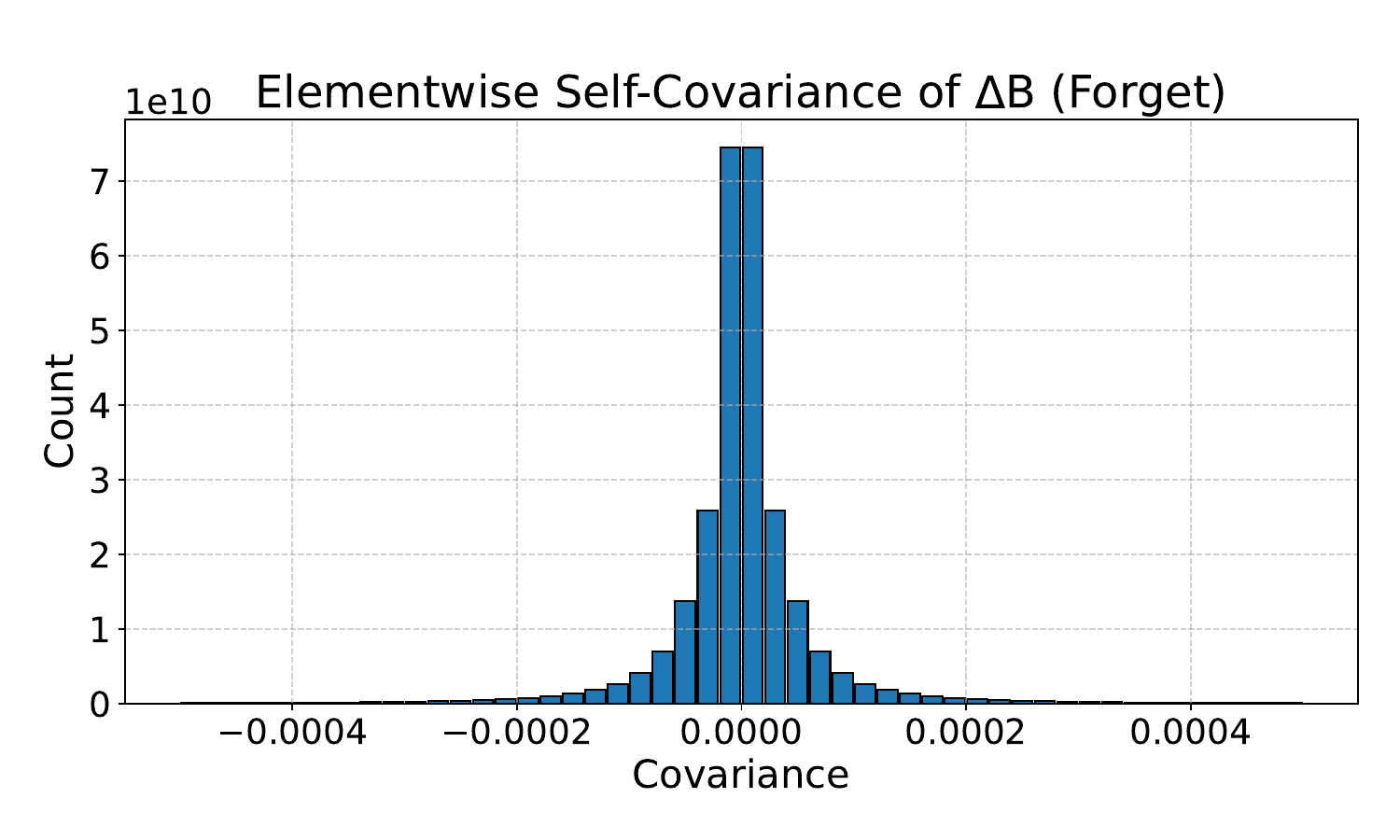}
        \caption{\(\Delta B\) on Forget set}
    \end{subfigure}
    \hfill
    \begin{subfigure}[b]{0.48\linewidth}
        \centering
        \includegraphics[width=\linewidth]{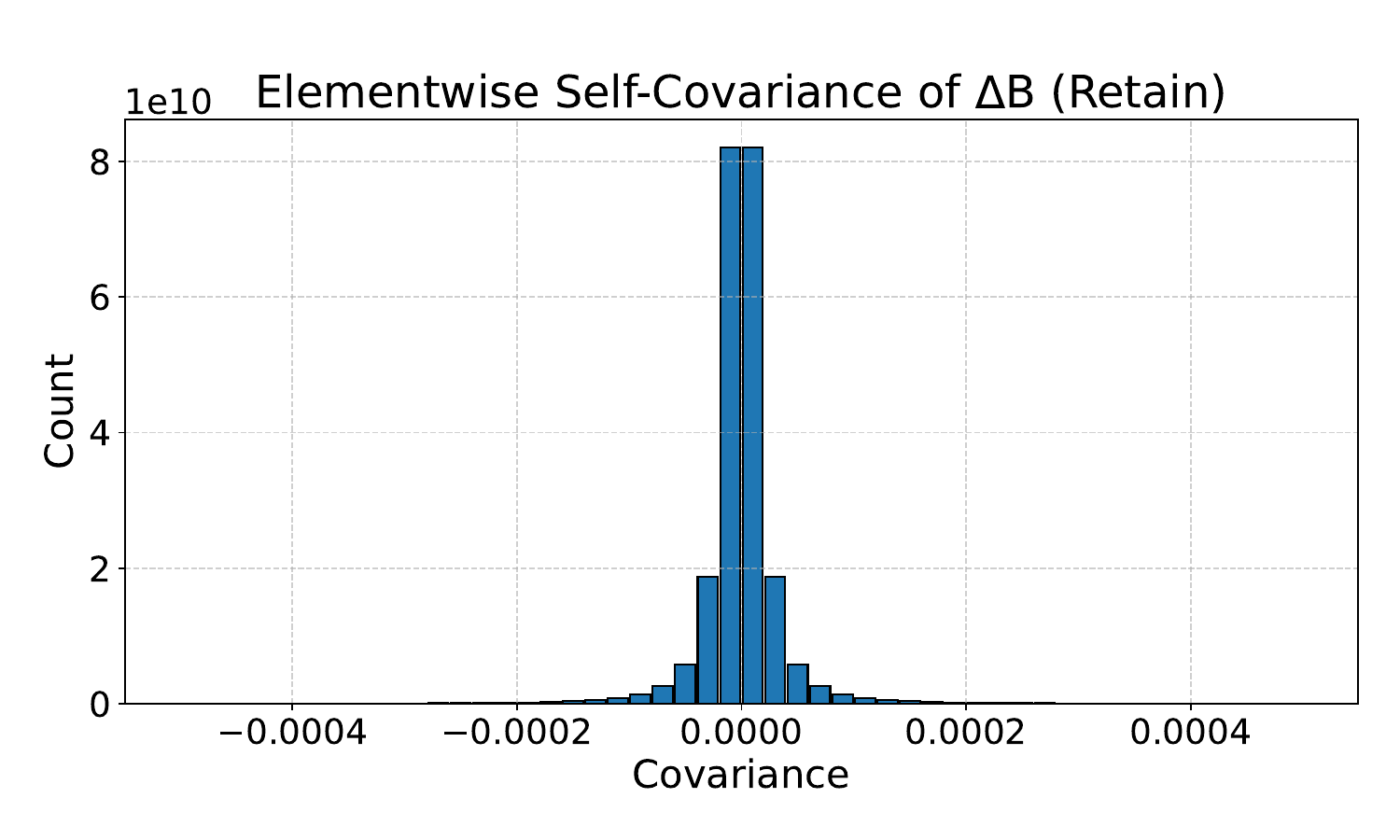}
        \caption{\(\Delta B\) on Retain set}
    \end{subfigure}

    \vspace{0.5em} 

    \begin{subfigure}[b]{0.48\linewidth}
        \centering
        \includegraphics[width=\linewidth]{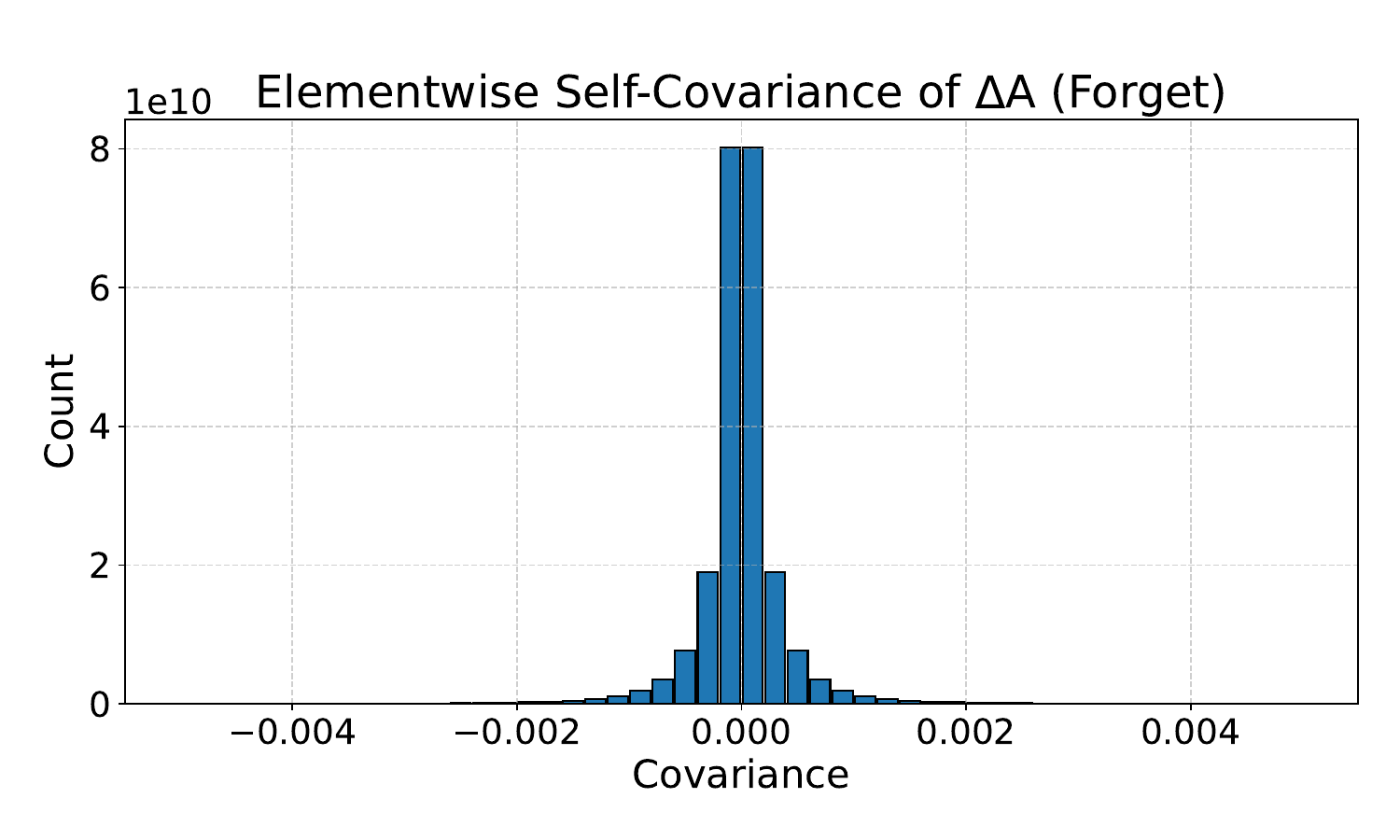}
        \caption{\(\Delta A\) on Forget set}
    \end{subfigure}
    \hfill
    \begin{subfigure}[b]{0.48\linewidth}
        \centering
        \includegraphics[width=\linewidth]{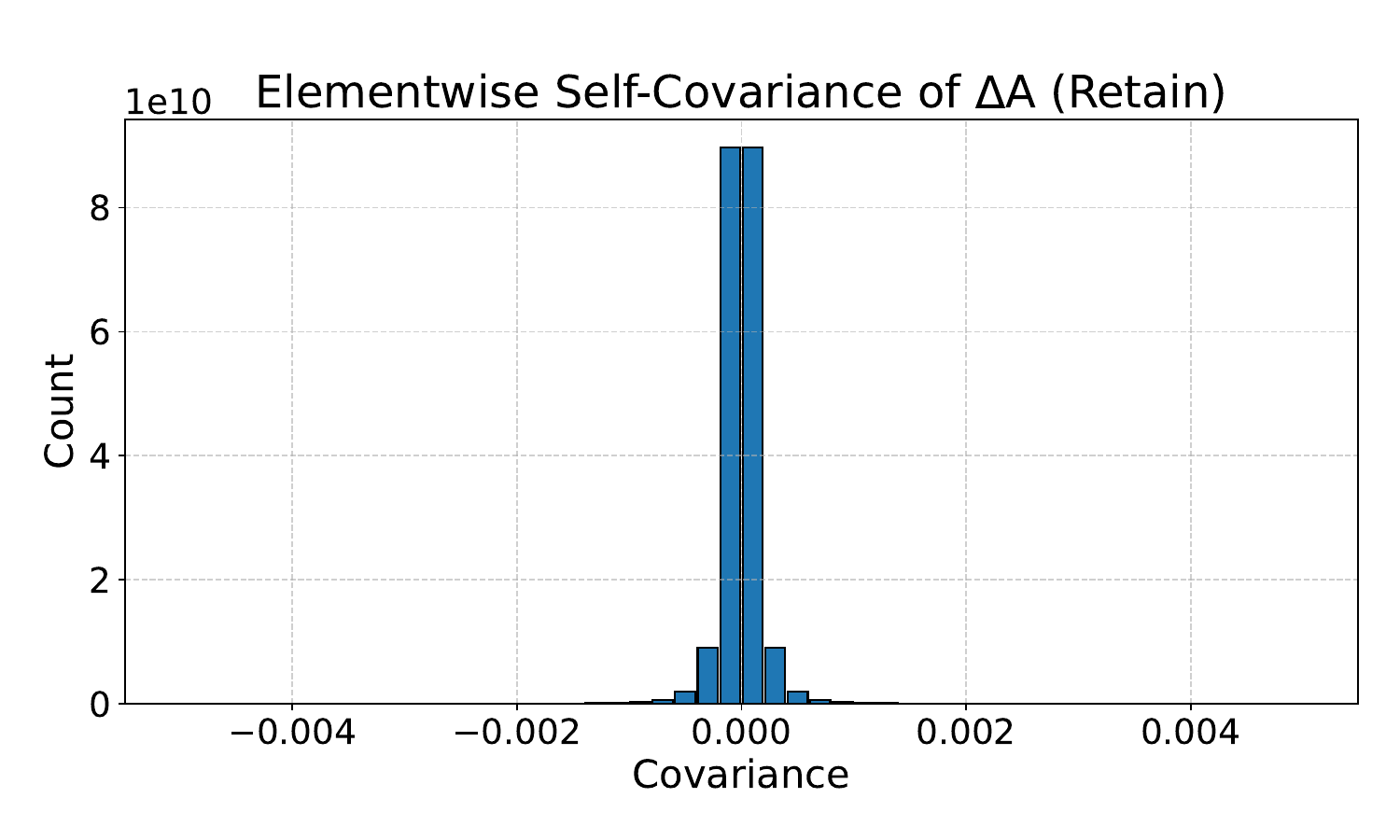}
        \caption{\(\Delta A\) on Retain set}
    \end{subfigure}

    \caption{\textbf{Element-wise self-covariance histograms of $\Delta A$ and $\Delta B$.} The gradients are computed separately over the forget and retain sets. Each plot illustrates the distribution of self-covariance values within the gradient matrices, with diagonal entries excluded from the counting to emphasize inter-parameter interactions.}
    \label{fig:Figure5_self_cov_histogram}
\end{figure}

As shown in Figure~\ref{fig:Figure5_self_cov_histogram}, the off-diagonal covariance terms in both $\Delta A$ and $\Delta B$ are sharply concentrated near zero for both the forget and retain sets. This empirically supports the approximation that distinct elements within each update matrix can be treated as independent. Therefore, the full factorization of the fourth-order expectation becomes practically valid, and the assumption of intra-matrix independence is well justified in our analysis.

\subsection*{A.4 Negligible expectation values of parameter updates}

We empirically observe that the squared expectation values $E_{\mathcal{D}}[\Delta B_{ik}]^2$ and $E_{\mathcal{D}}[\Delta A_{kj}]^2$ are relatively small (on the order of $10^{-1}$ to $10^{-2}$), although not strictly negligible.

We acknowledge that this assumption of negligible squared expectations is not strictly accurate; however, we made this simplifying assumption intentionally to achieve substantial computational efficiency. Specifically, this design choice enabled approximately a 100-fold reduction in memory consumption and a roughly 40-fold speed-up compared to previous approaches.

Our empirical analysis indicates that this approximation does not significantly degrade the method’s overall performance. Relaxing this assumption could potentially lead to further performance improvements; exploring such refinements will be an interesting direction for future work.

\subsection*{Implementation Remark}

Although our theoretical analysis assumes negligible squared expectations for simplicity, we observed empirically that the exact computation of variances for both $\Delta A$ and $\Delta B$, including their squared expectation terms, leads to significantly improved performance:
\[
\mathrm{Var}_{\mathcal{D}}[\Delta A_{kj}] = E_{\mathcal{D}}[(\Delta A_{kj})^2] - (E_{\mathcal{D}}[\Delta A_{kj}])^2,\quad
\mathrm{Var}_{\mathcal{D}}[\Delta B_{ik}] = E_{\mathcal{D}}[(\Delta B_{ik})^2] - (E_{\mathcal{D}}[\Delta B_{ik}])^2.
\]

This indicates that while the assumption \textbf{A.4} aids analytical simplicity and clarity, practical scenarios require careful consideration of nonzero expectation values for both $\Delta A$ and $\Delta B$ to achieve optimal performance.

\section{In-domain Unlearning Scenario}

In our main experiments, existing unlearning benchmarks typically assume the forget subset ($\mathcal{D}_f$) differs significantly from the retain subset. However, the general definition of unlearning does not inherently impose this assumption, implying the forget set could, in theory, be arbitrarily selected from the same distribution as the entire dataset. Thus, we conduct additional experiments explicitly designed to examine an \textit{in-domain unlearning scenario}, where the distribution of the forget set closely matches that of the overall dataset.

In practice, exactly reproducing an arbitrary forget subset scenario would require access to the entire pretraining corpus of the language model, which is typically unavailable. Therefore, we approximate the in-domain setting by randomly selecting 10\% of the question-answer pairs from the TOFU dataset as the forget set, ensuring its distribution aligns closely with the full dataset.

Because our forget set is randomly sampled, many forgotten Q\&A pairs lack perturbed answers, preventing direct computation of metrics such as the Truth Ratio. To overcome this limitation, we evaluate both Model Utility (MU) and Forget Quality (FQ) using ROUGE scores and probability-based metrics derived from model outputs. Model selection is based on evaluating MU and FQ through the harmonic mean of ROUGE and probability scores. We perform hyperparameter tuning over 15 trials and select the configuration that minimizes FQ while maintaining at least 95\% of the original MU. To ensure a fair comparison between FILA and our proposed method (VILA), we compute the importance map without the LoRA approximation, applying only the Fisher Information correction. Models used in this experiment include Phi-1.5B and Llama2-7B.

\begin{table}[htbp]
\centering
\vspace{0.5em}
\begin{tabular}{lcccc}
\toprule
Method & Llama2-7B MU $\uparrow$ & Llama2-7B FQ $\downarrow$ & Phi-1.5B MU $\uparrow$ & Phi-1.5B FQ $\downarrow$ \\
\midrule
IHL (Baseline) & 0.95 & 0.65 & 0.88 & 0.69 \\
IHL + FILA & 0.93 & {0.50} & 0.89 & {0.55} \\
IHL + Ours & 0.94 & 0.52 & 0.88 & 0.57 \\
\bottomrule
\end{tabular}
\caption{In-domain unlearning performance on the TOFU dataset. MU: Model Utility (higher is better), FQ: Forget Quality (lower is better).}
\label{tab:in_domain_results}
\end{table}

Table~\ref{tab:in_domain_results} summarizes our experimental results for the in-domain scenario under the IHL loss setting. The experimental results indicate that under the in-domain scenario, FILA and our method (VILA) exhibit relatively similar performance, with only minor differences. This aligns with theoretical expectations: the primary advantage of VILA arises when there is a significant distribution mismatch between the forget set and the overall dataset. When the forget set closely mirrors the overall dataset, the benefit of Fisher Information correction naturally diminishes, leading to comparable performance between FILA and VILA.

We further emphasize that, in realistic applications, the forget set typically does not represent the full data distribution and often significantly differs from it. Therefore, methods explicitly accounting for distributional differences between the forget set and the entire dataset, such as VILA, can provide substantial practical advantages.

\section{Validity of the Expectation as an Importance Score for Llama2-7B}\label{appendix:A_3_validity_of}

We conduct the same experiment described in Section~\ref{subsec:6_2_validity_of_the_expectation} on Llama2-7B. Table~\ref{tab:llama2-full-results-remove-utility} presents the results. 
\textsc{ExpILA} and \textsc{AbsILA} generally achieve comparable or better performance compared to FILA, particularly when the forget set size is small (Forget 1\%). As the forget set size grows (Forget 5\% and 10\%), \textbf{VILA} generally demonstrates better or competitive performance compared to \textsc{ExpILA} and \textsc{AbsILA}.
Importantly, these results empirically confirm our claim: when the distribution of the forget set differs from that of the entire data, the expectation values of the score function become different from zero.
Thus, correcting the original Fisher Information estimation, as done in VILA, becomes essential for robust and accurate importance estimation.

\begin{table}[t]
\centering

\begin{tabular}{l|l|ccc|c}
\toprule
\small
Loss & Method & Forget 1\% & Forget 5\% & Forget 10\% & AVG Gain ($\uparrow$) \\
\midrule
\multirow{4}{*}{GD} 
  & FILA     & -3.30 & -12.53 & -17.27 & -- \\
  & ExpILA   & -2.90 & -12.18 & -6.84 & 2.64 \\
  & AbsILA   & -2.90 & -9.02 & -9.06 & 2.95 \\
\rowcolor{gray!10}
  \cellcolor{white} & VILA & {-2.17} & {-1.40} & {-1.18} & \textbf{8.36} \\
\midrule
\multirow{4}{*}{NPO} 
  & FILA     & -3.30 & -11.18 & -11.06 & -- \\
  & ExpILA   & {-1.27} & -12.18 & -5.48 & 4.72 \\
  & AbsILA   & -2.52 & -12.18 & {-4.26} & 3.92 \\
\rowcolor{gray!10}
  \cellcolor{white} & VILA & -1.54 & {-4.32} & -4.59 & \textbf{5.74} \\
\midrule
\multirow{4}{*}{IHL} 
  & FILA     & -3.30 & -0.95 & -0.47 & -- \\
  & ExpILA   & -1.27 & -0.10 & -0.34 & 7.27 \\
  & AbsILA   & {-0.78} & {-0.01} & {-0.23} & \textbf{7.50} \\
\rowcolor{gray!10}
  \cellcolor{white} & VILA & -1.27 & -0.20 & -0.40 & 7.22 \\
\bottomrule
\end{tabular}
\caption{\textbf{Validity of the Expectation as an Importance Score.} Results using Llama2-7B model.}
\label{tab:llama2-full-results-remove-utility}
\end{table}

\section{LoRA Initialization Sensitivity (Sigma Ablation)}\label{sec:AppD_Sigma Ablation}

Purely zero-initialized LoRA parameters ($A=0, B=0$) yield zero gradients, making meaningful importance estimation impossible. Thus, appropriate initialization of LoRA parameters is critical. To empirically investigate the impact of initialization, we systematically evaluate various initialization strategies by varying the standard deviation ($\sigma$) from $0.01$ to $0.50$. 

\begin{table}[htbp]
\centering
\vspace{0.5em}
\begin{tabular}{lccccccc}
\toprule
Method & $\sigma=0.01$ & $\sigma=0.05$ & $\sigma=0.10$ & $\sigma=0.20$ & $\sigma=0.30$ & $\sigma=0.40$ & $\sigma=0.50$ \\
\midrule
GD   & $-13.54$ & $-12.41$ & ${-8.59}$ & $-12.13$ & $-10.29$ & X & X \\
NPO  & X        & $-10.80$ & $-11.32$ & ${-9.30}$  & $-9.79$  & X & X \\
IHL  & X        & ${-2.02}$  & $-9.54$  & $-11.06$ & $-10.54$ & X & X \\
\bottomrule
\end{tabular}\
\caption{Sensitivity of unlearning performance to LoRA initialization standard deviation ($\sigma$). Results show Forget Quality scores (lower is better) for Phi-1.5B model. ``X'' indicates unstable training or divergence. Best results per loss function are in bold.}
\label{tab:sigma_ablation_phi}
\end{table}

Table~\ref{tab:sigma_ablation_phi} summarizes our experimental results for the Phi-1.5B model across three unlearning loss functions: GD, NPO, and IHL. We observe that overly small initialization values (e.g., $\sigma=0.01$) led to unstable gradients, causing either suboptimal or failed unlearning performance (marked as ``X''). Similarly, overly large initialization values ($\sigma \geq 0.40$) introduce excessive noise, violating Assumption~A.1 (negligible magnitude of initial matrices) and resulting in unstable training or divergence.

We treat LoRA initialization as a hyperparameter and tune it equally across all methods using 15 validation trials. We confirm that effective and stable initialization values (e.g., $\sigma=0.05$) are easily identifiable, ensuring consistent and reliable importance estimation across experiments.

\section{Sensitivity to the Extent of Importance Map Usage}

To investigate how the performance of VILA is affected by the extent to which the importance map is applied, we vary the proportion of layers that receive the calculated importance map. We first compute the average importance score per layer, then selectively apply the importance map to the top $n\%$ layers ($n \in \{25, 50, 75\}$) ranked by these scores. The remaining layers are uniformly updated without importance weighting.

\begin{table}[htbp]
\centering
\vspace{0.5em}
\begin{tabular}{lccccc}
\toprule
Method & 0\% (Baseline) & 25\% Layers & 50\% Layers & 75\% Layers & 100\% (VILA) \\
\midrule
GD + VILA  & -16.61 & \textbf{-0.23} & -0.83 & -0.47 & -1.18 \\
IHL + VILA & -7.70  & \textbf{-0.01} & -0.03 & -0.29 & -0.40 \\
NPO + VILA & -13.84 & \textbf{-3.94} & -4.76 & -5.29 & -4.59 \\
\bottomrule
\end{tabular}
\caption{Ablation study on the sensitivity of VILA to the extent of importance map application. Results show Forget Quality scores (lower is better). Best performance per method is shown in bold.}
\label{tab:importance_map_extent}
\end{table}

Table~\ref{tab:importance_map_extent} summarizes the experimental results obtained using the TOFU benchmark with the Forget 10\% setting and Llama2-7B model. From these results, we make the following key observations:

\begin{itemize}
    \item Applying the importance map selectively to only the top 25\% of layers yields better unlearning performance compared to applying it to all layers (100\%).
    \item This suggests that VILA effectively identifies a subset of layers most critical to the unlearning process, resulting in improved forgetting performance when updates are focused on fewer, more relevant layers.
    \item Restricting updates to this smaller set of layers potentially reduces unnecessary parameter changes, thereby preserving model utility and achieving more targeted and efficient unlearning.
\end{itemize}

\section{Qualitative Analysis}

\begin{figure}[t]
  \centering
  \includegraphics[width=\textwidth]{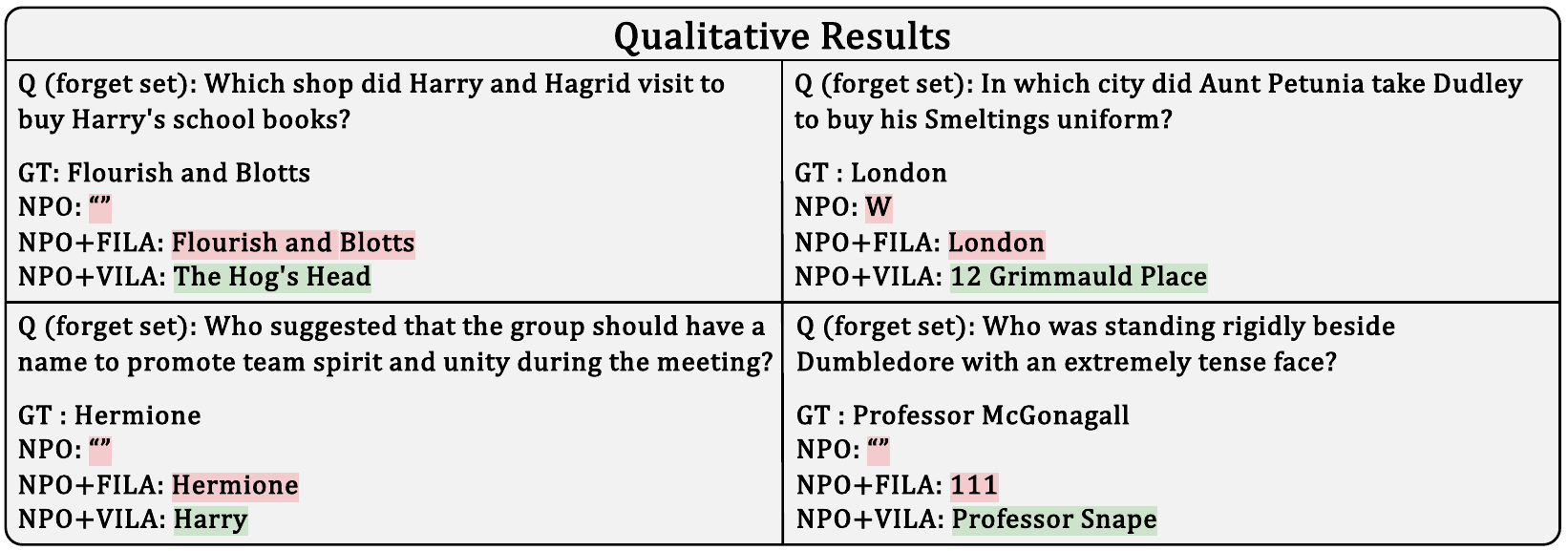}
  \caption{Qualitative examples from the MUSE Forget Knowledge benchmark. Red indicates responses that are either linguistically inconsistent or include the ground truth.
Green denotes plausible answers to the question that differ from the ground truth.}
  \label{fig:qualitative_results}
\end{figure}

Figure~\ref{fig:qualitative_results} shows response examples of each method on the MUSE forget set. Among the tested loss functions, NPO consistently demonstrates superior performance on the MUSE benchmark. Therefore, we conduct this qualitative analysis based on NPO and the cases where FILA and Ours are applied to NPO.

Our analysis reveals that although NPO successfully removes target information, its outputs often lack meaningful content. In many cases, the responses consist of blank spaces, punctuation, or other non-informative symbols. While this technically satisfies the objective of removing correct answers, it fails to generate linguistically coherent or contextually appropriate text, limiting its practical utility. Furthermore, combining NPO with FILA frequently results in the reproduction of correct answers, indicating ineffective unlearning. In contrast, our method not only avoids the target information but also maintains fluency and semantic coherence, demonstrating both high benchmark performance and practical unlearning effectiveness.

\section{Forget Performance Trajectory}

Reporting the trajectories of forgetting performance throughout the unlearning process provides valuable insights into the effectiveness and stability of unlearning methods. To comprehensively evaluate our method (VILA), we conduct additional experiments on the TOFU benchmark, explicitly tracking the forget quality and model utility at each step of the unlearning training process.

\begin{figure}[htbp]
  \centering

  \begin{subfigure}[b]{0.95\textwidth}
    \centering
    \includegraphics[width=\linewidth]{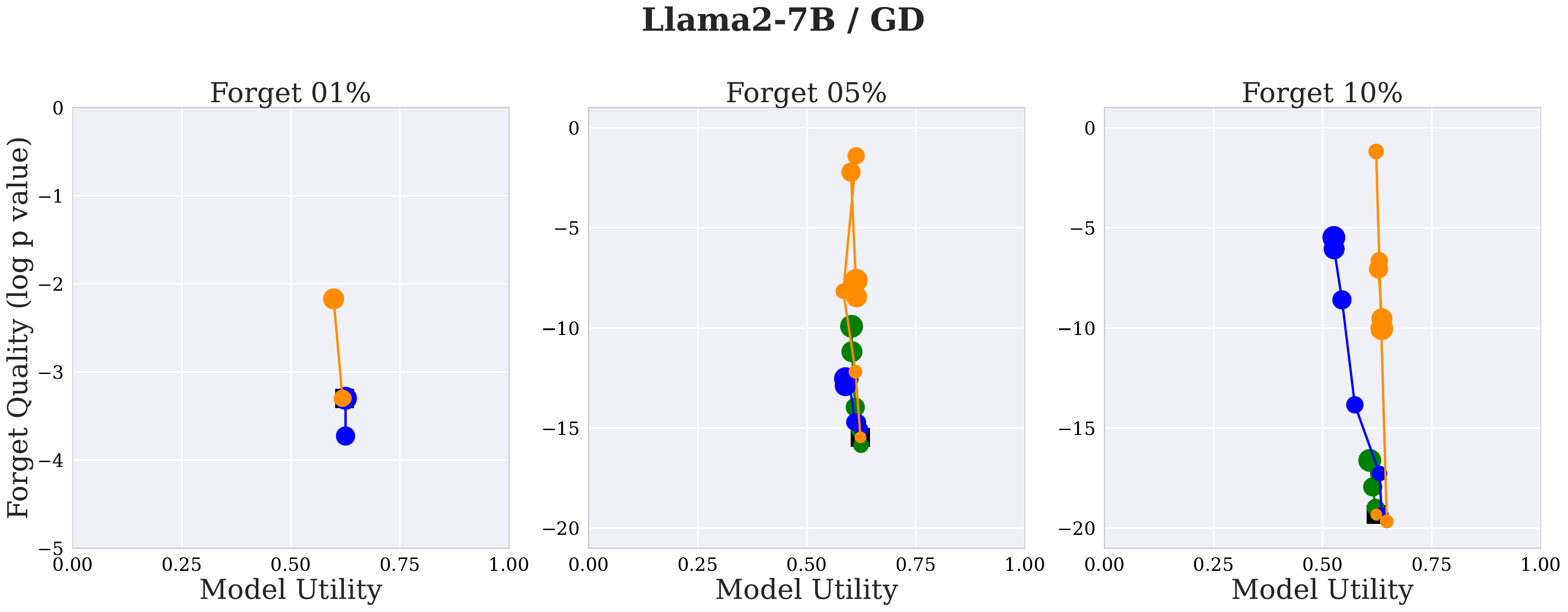}
    \label{fig:gd_unlearning}
  \end{subfigure}

  \vspace{1em}  

  \begin{subfigure}[b]{0.95\textwidth}
    \centering
    \includegraphics[width=\linewidth]{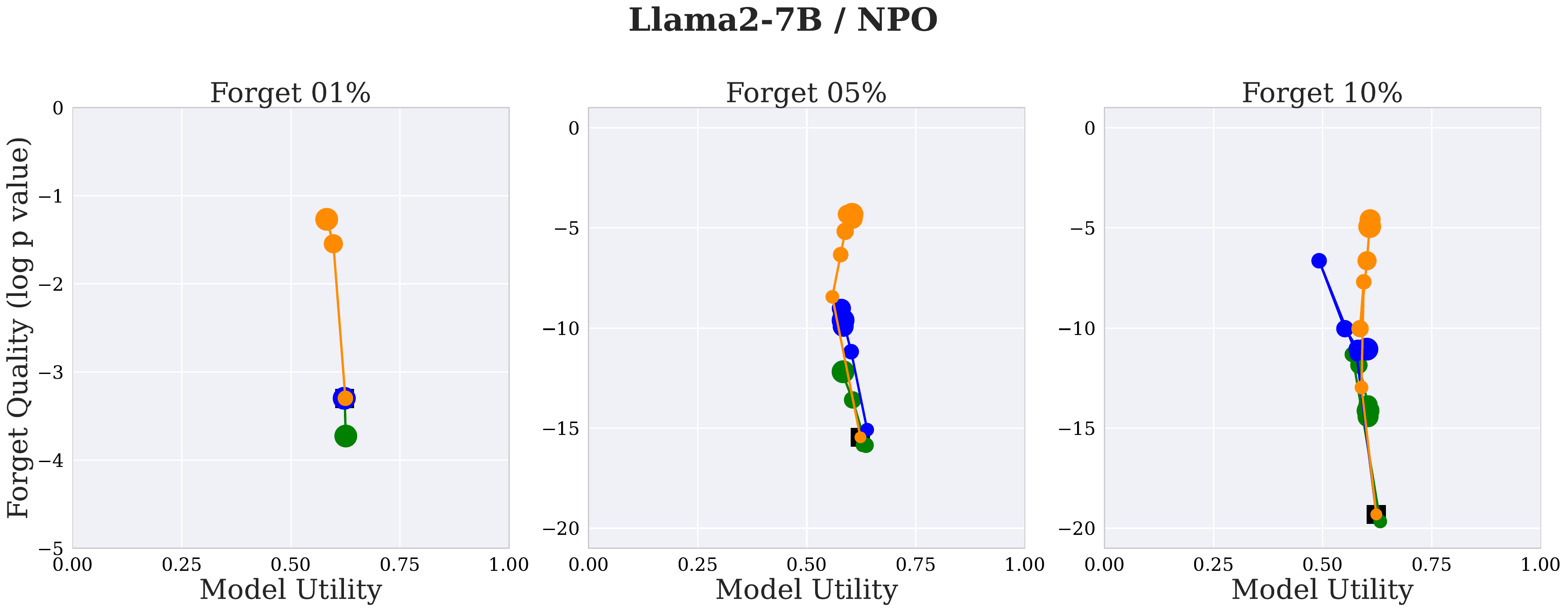}
    \label{fig:npo_unlearning}
  \end{subfigure}

  \vspace{1em}

  \begin{subfigure}[b]{0.95\textwidth}
    \centering
    \includegraphics[width=\linewidth]{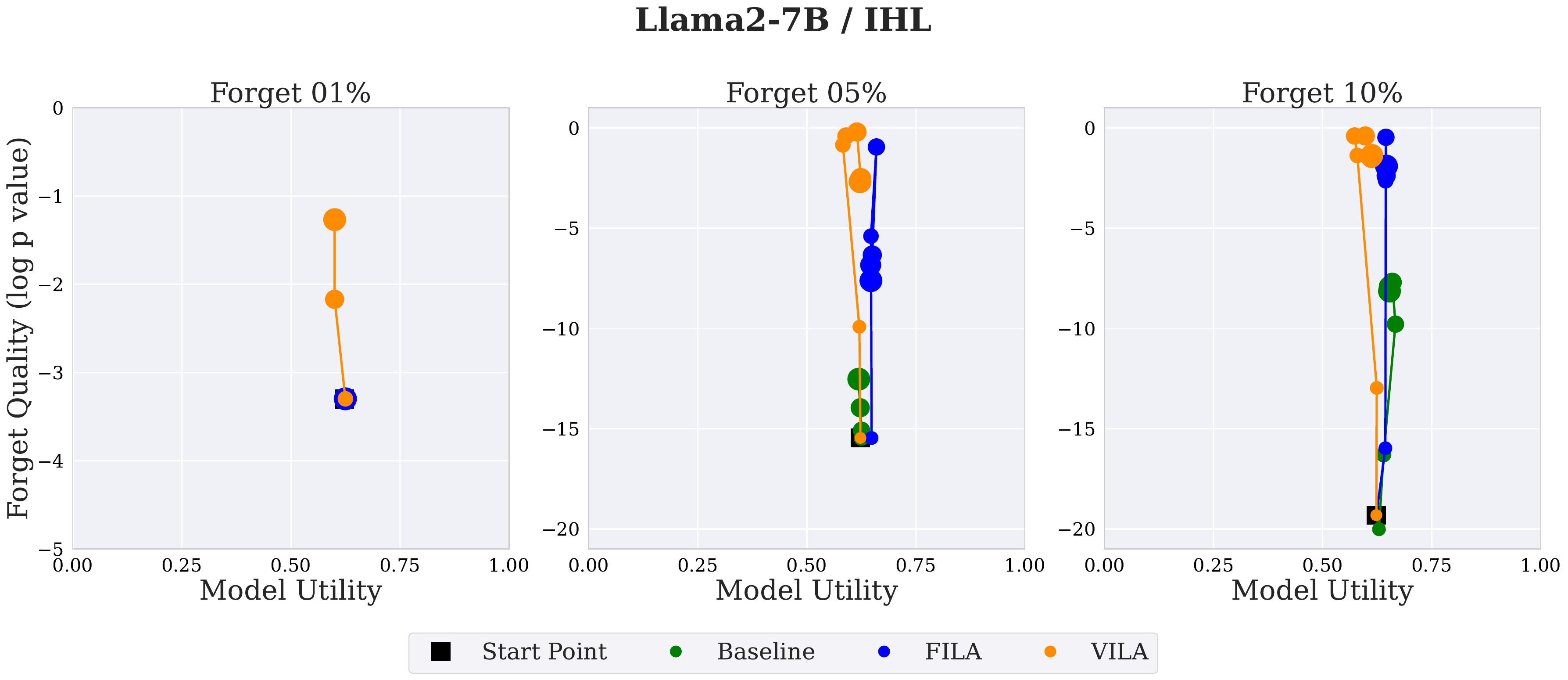}
    \label{fig:ihl_unlearning}
  \end{subfigure}

  \caption{Llama2-7B Unlearning Trajectories.}
  \label{fig:llama2_tracjectory}
\end{figure}

The experimental results demonstrate that VILA consistently achieves comparable or improved forgetting performance compared to FILA while notably preserving higher model utility. Notably, FILA achieves similar forget performance to VILA when paired with the IHL loss; however, VILA shows a consistent advantage in balancing forget performance with minimal degradation of model utility across various settings. Thus, rather than demonstrating absolute superiority in forget quality alone, VILA effectively provides a more favorable and stable trade-off throughout the entire unlearning trajectory.

\section{Benchmarks}
\label{asec:benchmarks}

\textbf{TOFU}~\citep{maini2024tofu} is a synthetic dataset consisting of 20 question–answer pairs for each of 200 fictional authors. The primary task is to effectively unlearn information corresponding to 1\%, 5\%, or 10\% of the total authors. The reference model is trained solely on the retain set (i.e., the remaining 99\%, 95\%, or 90\% of authors, respectively), serving as the oracle model that represents the ideal outcome of successful unlearning. Unlearning quality is measured by Forget Quality, defined as the p-value from the Kolmogorov–Smirnov test comparing the output distributions of the unlearned model and the reference model. A higher p-value indicates greater similarity between the two models’ outputs, suggesting more effective unlearning. Meanwhile, Model Utility evaluates how well the unlearned model preserves its performance on data excluding the forget set, assessing both accuracy on the retain set and general knowledge including factual information about real-world authors and commonsense reasoning. In our experiments, we follow the original TOFU setup and evaluate our method using both the Phi-1.5B and Llama2-7B backbone models.  

\textbf{WMDP}~\citep{li2024wmdp} is a multiple-choice benchmark designed to evaluate a model’s ability to unlearn knowledge related to hazardous domains such as biosecurity and cybersecurity. The forget set consists of scientific papers related to biosecurity and GitHub passages related to cybersecurity, while the retain set is composed of passages from Wikitext. Unlearning performance is assessed based on two criteria: lower accuracy on WMDP QA tasks indicates more effective forgetting of hazardous knowledge, while higher accuracy on general evaluation benchmarks such as MMLU~\citep{hendrycks2021mmlu} indicates better preservation of the model’s general capabilities. Following the original WMDP paper, we conduct experiments using the Zephyr-7B-\(\beta\)~\citep{tunstall2024zephyr} model.  

\textbf{MUSE Books}~\citep{shi2025muse} is an unlearning benchmark constructed from the Harry Potter book series~\citep{rowling1997-2007harrypotter}. It evaluates unlearning performance through two complementary metrics: verbatim memorization (VerbMem) and knowledge-based generation (KnowMem). VerbMem measures whether the model has successfully forgotten specific content from the forget set by calculating the ROUGE-L F1 score~\citep{lin-2004-rouge} between the model’s output and the original data. On the other hand, KnowMem evaluates whether the model can still generate correct answers when given question–answer pairs from the dataset. It is computed as the average ROUGE score~\citep{lin-2004-rouge} between the model’s output and the ground-truth answer, and is used to assess performance on both the forget and retain sets. Following the original MUSE paper, we use ICLM-7B~\citep{shi2024incontext} as the backbone language model. According to the original MUSE setup, lower VerbMem and KnowMem scores on the forget set indicate better unlearning, while higher KnowMem scores on the retain set indicate better preservation of relevant knowledge. However, we argue that this setting is unrealistic for practical applications, and we discuss our rationale in detail in Section~\ref{asec:exp_designs}.

\section{Compared Methods}
\label{asec:compared_methods}

\textbf{Gradient Difference (GD)}~\citep{liu2022continual} applies the Gradient Ascent (GA) loss to the forget set \( D_f \) and the standard negative log-likelihood loss to the retain set, as follows:
\begin{equation}
\label{eq:ga_loss}
\mathcal{L}_{\text{GD}}(\theta) = 
- \mathbb{E}_{(x,y) \sim \mathcal{D}_f} 
\left[ -\log \left( p(y \mid x; \theta) \right) \right]
+ \mathbb{E}_{(x,y) \sim \mathcal{D}_r} 
\left[ -\log \left( p(y \mid x; \theta) \right) \right].
\end{equation}
GA intentionally maximizes the prediction loss on the forget data which lowers the generation probability of the forget tokens and consequently discourages the model from producing them.

\textbf{Negative Preference Optimization (NPO)}~\citep{zhang2024negative} extends the concept of preference optimization—originally designed to train models to favor more desirable responses—to the unlearning setting. Specifically, NPO treats the responses in the forget set \( \mathcal{D}_f\) as negative examples and adjusts the model to minimize their selection probability, while applying the standard negative log-likelihood loss to the retain set \( \mathcal{D}_r\), as follows: 
\begin{equation}
\label{eq:npo_loss}
\mathcal{L}_{\text{NPO}}(\theta) = 
- \frac{2}{\beta} \mathbb{E}_{(x, y) \sim \mathcal{D}_f} 
\left[ 
\log \sigma \left( 
- \beta \log \frac{p(y \mid x; \theta)}{p(y \mid x; \theta_{\text{ref}})} 
\right) 
\right]
+ \mathbb{E}_{(x,y) \sim \mathcal{D}_r} 
\left[ -\log \left( p(y \mid x; \theta) \right) \right].
\end{equation}Compared to GA, NPO provides a more stable gradient magnitude and mitigates the risk of catastrophic degradation in overall model performance caused by excessive loss on the forget set. 
This allows the model to better balance forget quality and overall utility.

\textbf{Inverted Hinge Loss (IHL)}~\citep{cha2025fila} is a loss function designed to address the limitations of GA. It decreases the predicted probability of the forget token while simultaneously identifying the most probable alternative token—excluding the forget token—and guides the model to increase the predicted probability of that token, as follows:
\begin{equation}
\label{eq:ihl_loss}
\mathcal{L}_{\text{IHL}}(\theta) = 
- \mathbb{E}_{(x,y) \sim \mathcal{D}_f} \left[
1 + p_\theta(y \mid x; \theta) 
- \max_{v \ne y} \left( p_\theta(v \mid x; \theta) \right)
\right]
+ \mathbb{E}_{(x,y) \sim \mathcal{D}_r} 
\left[ -\log \left( p(y \mid x; \theta) \right) \right].
\end{equation}
For the retain set \( \mathcal{D}_r\), the standard negative log-likelihood loss is applied. This formulation suppresses undesirable tokens while reinforcing plausible replacements, thereby preserving the fluency of the language model.

\section{Comprehensive Experimental Designs}\label{asec:exp_designs}
To ensure a fair comparison, we conduct the same number of hyperparameter searches for every combination of unlearning loss and initialization strategy. When defining the search space, we exclude extreme learning rates—those that are too small to cause any learning progress, or too large, resulting in immediate collapse of model performance. Specifically, we consider a model to have collapsed if its accuracy drops to the level of random guessing: 0.0 for MUSE (a generation task) and 0.25 for WMDP (a four-choice multiple-choice task).

For TOFU, we perform 15 random search trials with the following hyperparameter ranges: learning rate in $[1\mathrm{e}{-6}, 2\mathrm{e}{-4}]$, retain coefficient $\lambda$ in $[0.5, 2.0]$, and NPO-specific $\beta$ in $[0.01, 1.0]$. All methods are trained for five epochs, and evaluations are performed at every epoch, as the optimal stopping point may vary by method. For WMDP, we apply random search with learning rate in $[1\mathrm{e}{-7}, 2\mathrm{e}{-4}]$. Additionally, we search the retain coefficient $\lambda$ over $[0.5, 2.0]$ in increments of 0.1, and $\beta$ over $[0.01, 0.05]$ in increments of 0.01. All models are trained for up to 125 steps, and evaluation is performed every 25 steps. For MUSE, we conduct 10 random search trials with learning rate in $[1\mathrm{e}{-7}, 1\mathrm{e}{-4}]$, $\lambda$ in $[1, 10]$, and $\beta$ in $[0.05, 2.0]$. 
The performance of the original MUSE Books model was obtained from the results reported in the original MUSE paper.
Experiments are conducted over two epochs, with evaluation performed after each epoch, similar to TOFU.

Although prior work reports training MUSE models for up to 10 epochs \citep{shi2025muse}, we find this setting unrealistic. MUSE Books dataset comprises approximately 1.1M tokens in the forget set and 0.5M tokens in the retain set. As noted in~\citet{shi2025muse}, the retrain model is trained for five epochs, totaling 2.5M tokens. We argue that any unlearning method that requires more training than retraining contradicts the practical goal of unlearning. To ensure the practical efficiency of unlearning, we restrict the total training budget to two epochs. Furthermore, unlike the original benchmark, which measures unlearning performance by how close the model's performance is to zero, we instead assess the gap between the unlearned and retrain models. This is motivated by our observation that minimizing the VerbMem score on the forget set often leads to significant degradation of the model's language capabilities. Specifically, we find that when the VerbMem score on the forget set approaches zero, the model tends to produce abnormal behavior, such as failing to generate any response to a query or returning only meaningless tokens such as punctuation marks. Notably, even the retrain model achieve zero VerbMem score due to generalization effects. These observations suggest that using zero performance on the forget set as the sole target is neither realistic nor desirable.
Therefore, similar to the TOFU benchmark, we evaluate unlearning effectiveness by measuring how closely the outputs of the unlearned model align with those of the retrain model.

Finally, for each method we compare the unlearned model that achieves the best forgetting score while preserving the utility of the original language model.
In contrast to some prior work, we argue that comparing unlearning performance without controlling for utility can be misleading. A model may appear to perform well on the forget set, but if its language modeling ability is severely degraded, such comparison becomes meaningless.
To address this, we select the model that achieves the best forgetting score while maintaining at least 95\% of the original model's utility, following the evaluation protocol proposed in Task Arithmetic~\citep{ilharco2023editing}.
When multiple forgetting metrics are involved, we use the average score to select the best model.
Under this evaluation setting, model utility does not vary significantly across methods. Therefore, we report only forget set performance in all tables.

\section{Limitations and Future works}

Despite its strong empirical performance, our VILA has a limitation in that it requires both the forget set and the retain set for unlearning, which may restrict its applicability in real-world scenarios. As future work, we plan to develop methods that can perform unlearning without explicitly relying on a retain set.

\end{document}